\documentclass[lettersize,journal]{IEEEtran}
\usepackage{amssymb}
\usepackage{amsmath,amsfonts}
\usepackage{algorithmic}
\usepackage{algorithm}
\usepackage{array}
\usepackage[caption=false,font=normalsize,labelfont=sf,textfont=sf]{subfig}
\usepackage{textcomp}
\usepackage{stfloats}
\usepackage{url}
\usepackage{verbatim}
\usepackage{graphicx}
\usepackage{cite}

\usepackage{booktabs}
\usepackage{multirow}
\usepackage{hyperref}
\usepackage[table,xcdraw]{xcolor}
\usepackage{adjustbox}
\usepackage{xcolor}

\begin{document}

\title{LVMark: Robust Watermark for Latent Video Diffusion Models}

\author{Youngdong Jang, MinHyuk Jang, JaeHyeok Lee, Feng Yang,~\IEEEmembership{Senior Member,~IEEE,} \\ Gyeongrok Oh, Jongheon Jeong, Sangpil Kim
\thanks{Youngdong Jang, MinHyuk Jang, JaeHyeok Lee, Gyeongrok Oh, Jongheon Jeong, Sangpil Kim are with Department of Artificial Intelligence, Korea University, Seoul 02841, Republic of Korea (e-mail: altu1996@korea.ac.kr, wkdalsgur85@korea.ac.kr, dlwogurgur@korea.ac.kr, dhrudfhr98@korea.ac.kr, jonghj@korea.ac.kr, spk7@korea.ac.kr).}
\thanks{Feng Yang is with Google DeepMind, Mountain View, CA 94043, USA (e-mail: fengyang@google.com).}
\thanks{Youngdong Jang and MinHyuk Jang contributed equally to this work.}
\thanks{Corresponding author: Sangpil Kim}
}



\maketitle

\begin{abstract}
Rapid advancements in video diffusion models have enabled the creation of realistic videos, raising concerns about unauthorized use and driving the demand for techniques to protect model ownership. 
Existing watermarking methods suffer from two key limitations: they overlook temporal consistency due to conventional watermark decoders and degrade the visual quality of the generated videos.
To address these issues, we introduce a robust watermarking method for latent video diffusion models named Latent Video Diffusion Watermarking (LVMark).
We propose a novel watermark decoder tailored for generated videos by learning the consistency between adjacent frames.
It ensures accurate message decoding, even under malicious attacks, by combining the low-frequency components of the three-dimensional wavelet domain with the color features of the video.
Additionally, we train a latent decoder to maintain the visual fidelity of the generated video. Watermarks are embedded into layers with minimal impact on visual appearance using an importance-based weight modulation strategy.
We optimize both the watermark decoder and the latent decoder of diffusion model, effectively balancing the trade-off between visual quality and bit accuracy.
Our experiments show that our method embeds invisible watermarks into video diffusion models, ensuring robust decoding accuracy with 512-bit capacity, even under distortions.
\end{abstract}

\begin{IEEEkeywords}
Digital Watermarking, Copyright Protection, Diffusion Model
\end{IEEEkeywords}

\section{Introduction}
\label{sec:intro}
\IEEEPARstart{V}{ideo} diffusion models have made significant progress in generating high-quality videos~\cite{opensora, ma2024latte, blattmann2023stable, Chen_2024_CVPR, xing2025dynamicrafter}, leading to a growing supply of AI-generated video content. With these advancements, the need for methods to establish ownership of both the diffusion model and its creations has emerged. This issue can be addressed by embedding watermarks into the video diffusion model and enabling tracking. 

\begin{figure}[t!]
    \begin{center}
        \includegraphics[width=0.9\linewidth]{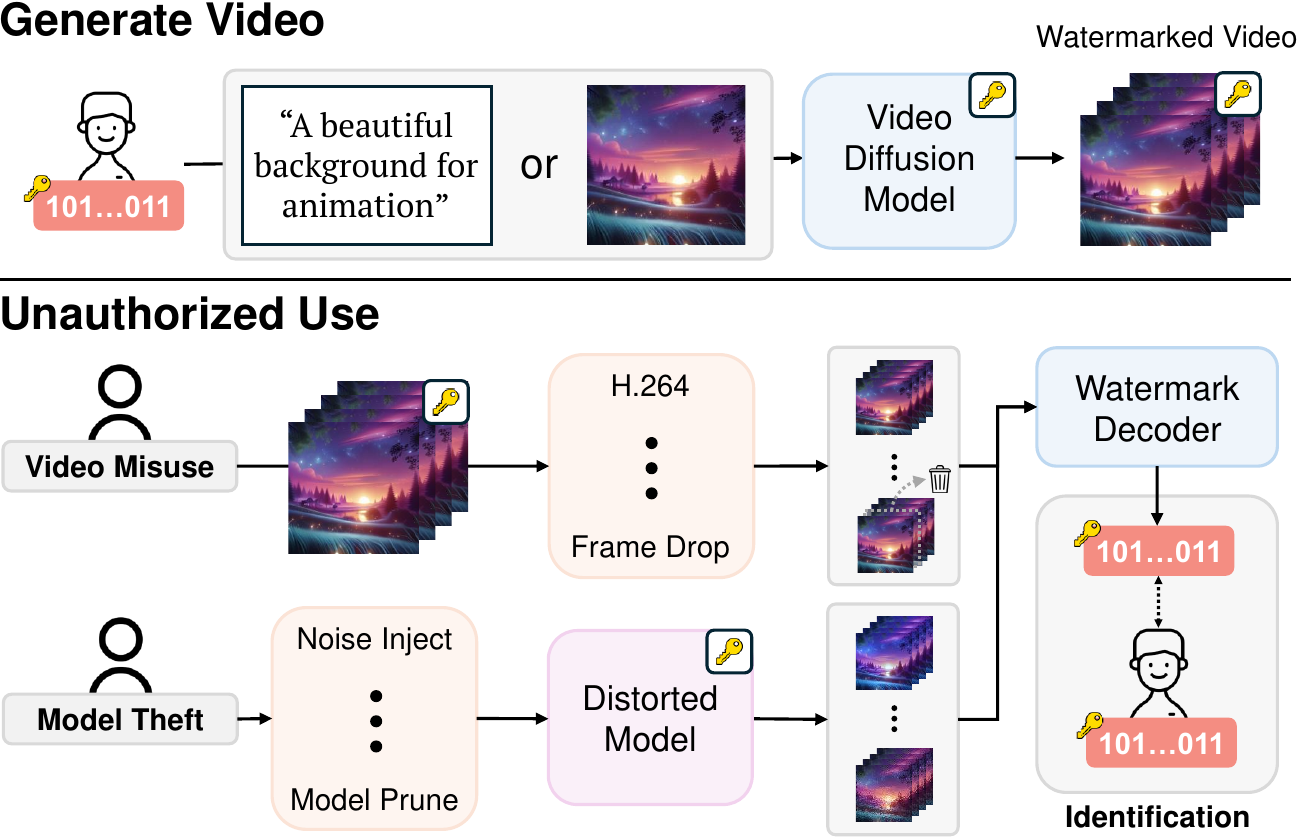}
    \end{center}
    \vspace{-1em}
    \caption{Overview of video generation and ownership identification. Top: An authorized user generates a watermarked video. Bottom: Despite distortions applied to the video or model, the watermark decoder can reliably identify the video's owner. }
    \label{fig:fig1}
\end{figure}

Existing watermarking methods for video diffusion model can be categorized into three types.
The first approach uses post-processing methods~\cite{zhu2018hidden, sander2024watermark, masoumi2013blind, zhang2023novel}. 
While these methods enable accurate watermark extraction, they can be easily bypassed if the model is stolen, as the watermark is not part of the model.
The second approach embeds watermarks into the latent noise of the diffusion model~\cite{wen2023treerings, hu2025videoshield, hu2025videomark}.
However, this approach generates altered content, and watermark extraction requires the inversion of the diffusion process, making it computationally intensive.
The third approach fine-tunes a pre-trained diffusion model to embed the watermark~\cite{fernandez2023stable, kim2024wouaf, huang2025video, liu2025implanting} ensuring watermark insertion cannot be bypassed.

To preserve the original content, the fine-tuning process keeps the latent noise of the diffusion model, while the latent decoder is trained to embed the watermark. These fine-tuning methods are categorized into two types. The first approach fine-tunes the latent decoder with the pre-trained watermark decoder. The primary drawback is that it requires a training process for each new watermark message. Alternatively, the second approach utilizes weight modulation, which directly modifies the model's parameters. This allows for the embedding of new watermarks without additional training once the model has been fine-tuned. However, the application of these fine-tuning methods to video diffusion models has not yet been researched in depth. While 2D watermark decoders can be applied to individual frames, they fail to preserve temporal consistency and leverage the large data volume of videos for high-capacity watermarking. 
Moreover, despite the emergence of Diffusion Transformer (DiT)~\cite{peebles2023scalable} enabling diffusion models to generate more diverse and high-quality content, existing watermarking methods have not yet been explored in this architecture. These underscore the need for a watermarking method compatible with DiT-based diffusion models and a video-specific decoder tailored for video diffusion models.

To tackle these challenges, we introduce Latent Video Diffusion Watermarking (LVMark), a fine-tuning-based watermarking framework for video diffusion models that is applicable to both U-Net~\cite{ronneberger2015u} and DiT~\cite{peebles2023scalable} architectures. 
Extracting watermarks solely from the RGB domain forces the model to embed them in visible color regions, which degrades video quality and makes the watermark vulnerable to distortions. 
Although the 3D wavelet transform (3D DWT) is widely used to enhance robustness and visual quality, we observe that watermark decoding based solely on 3D wavelet features becomes unstable in diffusion-based models.

Motivated by these observations, we integrate both RGB and 3D wavelet representations for watermark decoding. Specifically, watermark information is embedded in the frequency domain using the 3D DWT, while RGB features provide the primary spatial cues for reliable decoding. A cross-attention module~\cite{jaegle2021perceiver, jaegle2021perceiverio} effectively fuses these two representations, enabling robust watermark extraction.
Additionally, we use the low-frequency subband of the 3D wavelet domain, as video compression methods like H.264 tend to remove high-frequency regions.
The watermark decoder is jointly optimized with the latent decoder of the diffusion model by iteratively embedding and extracting random messages, enabling reliable bit extraction even under parameter variations. During the training process, we address the non-differentiable nature of H.264 compression by employing a small proxy network. This network approximates the complex compression operations, acting as a differentiable surrogate that significantly enhances the robustness of the watermark.

To embed random watermarks in the diffusion model, we utilize weight modulation~\cite{kim2024wouaf}, which directly modifies the model's parameters to embed the watermarks into the generated videos.
Prior fine-tuning methods~\cite{kim2024wouaf,fernandez2023stable} update all parameters of the model during fine-tuning. However, modulating all layers of the latent decoder disrupts temporal consistency and degrades video quality. Thus, we selectively modulate the latent decoder's layers with minimal perceptual impact to enhance invisibility. Selective parameter methods~\cite{foster2024fast, cai2025targeted} rely on analytical metrics like gradients or Fisher Information. However, by relying on localized derivatives and diagonal approximations~\cite{kirkpatrick2017overcoming,kunstner2019limitations}, these metrics inherently assume parameter independence and struggle to capture the complex spatio-temporal dependencies of video diffusion models.
Consequently, we utilize a perceptual metric aligned with human visual quality, namely LPIPS~\cite{zhang2018unreasonable}. Specifically, we evaluate the visual impact of each layer by injecting Gaussian noise into individual layers and measuring this metric. Furthermore, we propose a weighted patch loss that emphasizes regions prone to visual artifacts. The loss assigns larger weights to patches that are more susceptible to perceptual degradation, encouraging the model to suppress localized spatio-temporal artifacts introduced by watermark embedding. This design enables the model to maintain high-quality video generation while preserving reliable watermark decoding.

Experimental results show that our method outperforms all reasonable baselines, particularly in terms of video consistency as well as robustness against attacks. 
In particular, embedding 512-bit watermarks into Open-Sora~\cite{opensora} and DynamiCrafter~\cite{xing2025dynamicrafter} shows superior temporal consistency. Our method maintains high bit accuracy across both models even under rigorous distortions, such as H.264 compression, cropping, and frame-dropping.
We can establish model ownership even in cases of unauthorized use, as depicted in Figure~\ref{fig:fig1}, thanks to our method's ability to embed high-capacity, imperceptible watermarks into video diffusion models.

Our contributions can be summarized as follows:

\begin{itemize}
   \item We introduce LVMark, a robust watermarking method for video diffusion models that can be applied to both U-Net and DiT-based architectures.
   \item We propose a watermark decoder that leverages low-frequency of 3D wavelet components for robust spatio-temporal message decoding, ensuring resilience to video-specific attacks like H.264 compression and frame drops.
   \item By leveraging an importance-based weight modulation technique, our method embeds multiple messages into video diffusion models with a single training process while preserving video quality.
   \item To balance visual quality and message decoding accuracy, we introduce a weighted patch loss that effectively reduces localized artifacts.
\end{itemize}

\section{Related Work}
\label{sec:related_works}

\subsection{Video Diffusion Models}
Diffusion models have become a powerful generative approach, refining noisy data progressively to produce high-quality outputs~\cite{sohl2015deep, saharia2022palette, nichol2021improved, ho2022video, yin2024one}. 
Latent Video Diffusion Models (LVDMs)~\cite{blattmann2023align} optimize spatio-temporal features within a latent space, reducing computational costs. 
OpenSora~\cite{opensora} and Latte~\cite{ma2024latte} utilize the DiT~\cite{peebles2023scalable} to model long-range temporal dependencies. 
In contrast, Stable Video Diffusion~\cite{blattmann2023stable} and VideoCrafter~\cite{Chen_2024_CVPR} leverage U-Net~\cite{ronneberger2015u} to effectively capture both spatial and temporal dependencies. 
DynamiCrafter~\cite{xing2025dynamicrafter} enhances U-Net by modeling dynamic scenes and complex motion~\cite{yu2023video}. 
Both U-Net and DiT architectures play crucial roles in advancing video generation, highlighting the need for watermark methods that can be applied to both types of video diffusion models.

\subsection{Diffusion Model Watermark}
Digital watermarking~\cite{noorkami2008digital,asikuzzaman2014imperceptible,zhu2018hidden,luo2023dvmark, masoumi2013blind} plays a crucial role in protecting intellectual property, ensuring content authenticity, and enabling ownership tracking. 
Watermarking methods for diffusion models~\cite{wen2024tree, kim2024wouaf, fernandez2023stable, hu2025videoshield, hu2025videomark, liu2025implanting, huang2025video} embed watermarks into the generated content from diffusion models without a post-processing step.
Tree-Ring\cite{wen2024tree} is the first method to embed watermarks into the latent noise of a diffusion model. 
VideoShield~\cite{hu2025videoshield} and VideoMark~\cite{hu2025videomark} are also diffusion watermarking methods that embed the watermark into the latent noise. They modify the generated content and require inversion of the diffusion process for watermark decoding, making it computationally intensive.
In contrast, the fine-tuning method with a pre-trained watermark decoder preserves the generated content.
Stable Signature\cite{fernandez2023stable} is a fine-tuning method that embeds the watermark into the generated output.
Video Signature~\cite{huang2025video} introduces a Temporal Alignment module that guides the latent decoder to produce coherent frames during fine-tuning. Stable Video Signature~\cite{liu2025implanting} utilizes a 3DCNN to introduce temporal attention blocks to embed watermarks into the latent space. These methods require additional training to embed a new watermark once the model has been fine-tuned.
Separately, WOUAF~\cite{kim2024wouaf} is a fine-tuning method that modulates the weights of all layers, allowing for the embedding of new watermarks at inference time without requiring additional fine-tuning per message. However, like other existing methods~\cite{fernandez2023stable, huang2025video, liu2025implanting}, it typically relies on 2D watermark decoders, which limits its ability to capture temporal consistency across frames. Therefore, there is a crucial need for video watermarking methods that can preserve temporal consistency without degrading the generated content.

\begin{figure*}[t!]
    \begin{center}
        \includegraphics[width=1.0\textwidth ]{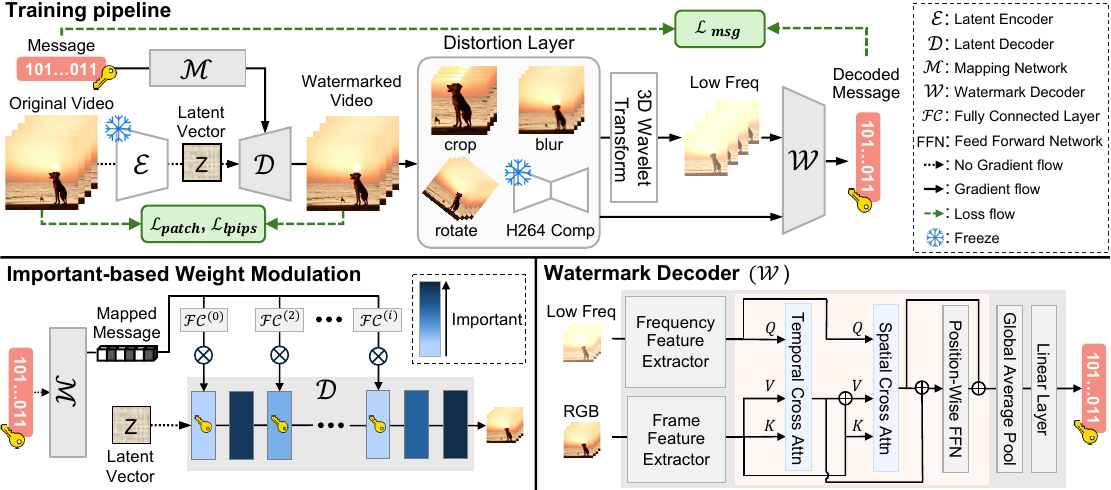}
    \end{center}
    \vspace{-1.5em}
    \caption{Training pipeline of LVMark. Top: We fine-tune the latent decoder to embed binary messages in generated videos and train the watermark decoder to extract the messages from distorted videos. We freeze the latent encoder $\mathcal{E}$ to preserve the original video content. At each iteration, the message is randomly sampled and transformed by the mapping network $\mathcal{M}$ into the input of a fully connected layer $\mathcal{FC}$. $\mathcal{FC}$ match the dimension to modulate each layer's weight parameters in the latent decoder. The modulated latent decoder decodes the latent vector $z$ into the watermarked video. This video is passed through a distortion layer that applies random distortion. To extract the watermark, we transform the distorted video into 3D wavelet subbands using the 3D Discrete Wavelet Transform (3D DWT). We utilize only the low-frequency subband (LLL) along with the RGB video as inputs to the watermark decoder $\mathcal{W}$, which finally extract the message.
    Bottom-left: We modulate layers of the latent decoder that minimally impact visual quality to embed random messages. Bottom-right: The watermark decoder combines the RGB video with low-frequency subbands from a 3D wavelet transform using cross-attention to decode the binary message.}
    \label{fig:fig2}
\end{figure*} 

\subsection{3D Discrete Wavelet Transform} Embedding watermarks in the frequency domain has been shown to improve robustness against distortions while enhancing invisibility. Consequently, numerous studies~\cite{larbi2018embedding, jimson2018dft, ko2020robust, gao2019dynamic,wang2017hybrid, wang2012informed, coria2008video} have explored frequency domain for effective watermarking. 

Among frequency domain techniques, 3D Discrete Wavelet Transform (3D DWT) is a classical method widely used to analyze temporal information of videos. 
3D DWT operates along the time, height, and width of videos to decompose a low-frequency subband and several high-frequency subbands: \(LLL\), \(LLH\), \(LHL\), \(LHH\), \(HLL\), \(HLH\), \(HHL\), and \(HHH\) each with a size of $F/2 \times H/2 \times W/2$, where \(L\) and \(H\) denote low and high-frequency, while $F$, $H$, and $W$ are frame, height, and width of video respectively. 
Existing video processing methods~\cite{zhang2014video, dharejo2022fuzzyact, yu2010video, mehrseresht2006efficient, yousefi2018novel} leverage the properties of 3D DWT to capture temporal information, while video digital watermarking techniques~\cite{al2009robust, liu2002robust, sakib2020robust, masoumi2013blind, li2007video} utilize 3D DWT to enhance the robustness of the watermark.
To the best of our knowledge, no research has yet explored watermarking video diffusion models using 3D DWT, underscoring the need for investigation to enhance robustness against common video modifications.

\section{Preliminaries.}
We utilize latent video diffusion models, which consist of a latent encoder \(\mathcal{E}\), a latent decoder \(\mathcal{D}\), and a diffusion model, which is typically based on one of two architectures: U-Net~\cite{ronneberger2015u} or DiT~\cite{peebles2023scalable}.
The latent encoder \(\mathcal{E}(v)\) maps the video \(v \in \mathbb{R}^{d_v}\) to a latent representation \(z \in \mathbb{R}^{d_z}\), where $d_v$ and $d_z$ denote the dimensions of video $v$ and latent vector $z$. 
In the forward diffusion process, a noise scheduler progressively adds Gaussian noise $\epsilon$ to the latent vector $z$ over $T$ timesteps. At a specific timestep $t$, the noisy latent $z_t$ is expressed as:
\vspace{-0.5em}
$$z_t = \sqrt{\bar{\alpha}_t} z + \sqrt{1 - \bar{\alpha}_t} \epsilon, \quad \epsilon \sim \mathcal{N}(0, I)$$, where $\bar{\alpha}_t$ is the noise schedule parameter at timestep $t$. During the backward process, the network trains to denoise $z_t$ to recover the original latent $z$, which is conditioned on auxiliary inputs $c$ (e.g., text prompts). The training objective of the latent diffusion model is typically formulated as minimizing the mean squared error between the added noise and the predicted noise:$$\mathcal{L}_{LDM} = \mathbb{E}_{z, c, \epsilon \sim \mathcal{N}(0, I), t} \left[ \left\| \epsilon - \epsilon_\theta(z_t, t, c) \right\|_2^2 \right]$$where $\epsilon_\theta$ denotes the noise prediction network. Once the denoising process is complete, the latent decoder $\mathcal{D}$ converts the latent representation back to the pixel space, yielding the reconstructed video $\hat{v} = \mathcal{D}(z)$.
U-Net-based diffusion models focus on local features through downsampling and upsampling layers.
In contrast, DiT-based models capture global dependencies and long-range correlations, making them effective for modeling complex temporal relationships. 

\section{Methods} 

An overview of our method is shown in Figure~\ref{fig:fig2}.
We introduce a novel watermark decoder that incorporates the video's temporal consistency (Section~\ref{subsec:Robust Video Watermark Decoder}).  
We selectively modulate the weights of the latent decoder $\mathcal{D}$ to embed imperceptible random watermarks into the model (Section~\ref{subsec:Importance-based Weight Modulation}).  
The distortion layer ensures robust watermark decoding against distortions like crop, rotation, and H.264 compression (Section~\ref{subsec:Distortion Layer}).  
By iteratively optimizing the watermark and latent decoders with random watermark messages, our method embeds an imperceptible watermark while ensuring robust extraction (Section~\ref{subsec:Training Objectives}).

\subsection{Robust Video Watermark Decoder}
\label{subsec:Robust Video Watermark Decoder}
We propose a new watermark decoder \( \mathcal{W} \) that extracts messages from generated videos by leveraging spatio-temporal information through the fusion of two domains: the generated RGB video \( \hat{v} \) and its 3D wavelet transformed subband \( \hat{v}_{freq} \), as depicted in the bottom-right of Figure~\ref{fig:fig2}.
We use only the low-frequency subband \(LLL\) among the 3D wavelet subbands to enhance robustness against video compression, as various compression methods discard high-frequency features~\cite{lee2012perceptual}. 

The watermark decoder is composed of two primary components: a feature extraction module and a feature fusion module. 
Before combining the RGB domain and frequency domain, we first extract each domain's features independently. Using six 3D convolutional blocks, we extract wavelet features \( F_{freq} \in \mathbb{R}^{B \times 2048 \times F \times H \times W} \) from the low-frequency subband \(LLL\), where $B$, $F$, $H$, and $W$ denote batch, frame, height, and width, respectively.  On the other hand, we employ the ResNet50 model to extract spatial features \( F_{rgb} \in \mathbb{R}^{B \times 2048 \times F \times H \times W} \) from the RGB video \( \hat{v} \).

To efficiently fuse these two features, we employ temporal and spatial cross-attention mechanisms. 
The temporal cross-attention measures the attention score across frames, while the spatial cross-attention measures the attention score across height and width. 
The temporal and spatial cross-attention modules share the same architecture, each designed with 8 multi-heads.
The entire feature fusion module consists of temporal cross-attention, spatial cross-attention, and a position-wise feed-forward network, with a depth of 2.
We set the wavelet features as the query \( Q = \text{FC}_q(F_{freq}) \), while using the spatial features as the key \( K = \text{FC}_k(F_{rgb}) \) and the value \( V = \text{FC}_v(F_{rgb}) \), where $\text{FC}_{q}$, $\text{FC}_{k}$ and $\text{FC}_{v}$ denote fully connected layers for query, key, and value, respectively. 
This allows the spatial feature \( F_{rgb} \) to capture information from the wavelet features \( F_{freq} \) by leveraging the attention scores computed between them. 
The cross-attention module $CA$ is defined as:
\begin{equation}
CA(F_{freq}, F_{rgb}) = \text{softmax}\left(\frac{Q \cdot K^T}{\sqrt{d_k}}\right) \cdot V.
\label{input:cross attention}
\end{equation}

The output of each cross-attention module is added to the spatial feature \( F_{rgb} \) for residual connection. 
With this watermark decoder, we obtain the message hidden within the generated video.

\subsection{Importance-based Weight Modulation}
\label{subsec:Importance-based Weight Modulation}
To embed random messages into the diffusion model, we propose an importance-based weight modulation module inspired by WOUAF~\cite{kim2024wouaf}, which modulates the weight parameters of the pre-trained latent decoder $\mathcal{D}$.
Although WOUAF successfully embeds messages into the model, our experiments show that modulating all layers degrades visual quality by altering parameters crucial for video generation.
To mitigate this, we propose a novel approach that modulates only a subset of layers with lower importance, as shown in the bottom-left of Figure~\ref{fig:fig2}.
As a pre-processing step, we quantify layer importance by sequentially perturbing each layer with noise sampled from a Gaussian distribution, \( p \sim \mathcal{N}(0, 0.04) \), resulting in perturbed weights: \( \mathcal{D}^{(i)}_{perturbed} = \mathcal{D}^{(i)} + p \).
By evaluating the LPIPS~\cite{zhang2018unreasonable} metric, which measures perceptual similarity, on videos generated with sequentially perturbed layers, our method identifies which layers have a greater impact on video quality.

In the training phase, we transform the random message $m$ into the mapped message $m'$ with the mapping network \( \mathcal{M}(m) : \mathbb{R}^{d_m} \rightarrow \mathbb{R}^{d_{m'}} \), which is composed of two fully connected layers. 
Then, we modulate the $i$-th selected layer $\mathcal{D}^{(i)}$ by performing element-wise multiplication between $\mathcal{D}^{(i)}$ and the mapped message, where the $i$-th fully connected layer \( \mathcal{FC}^{(i)}(m') : \mathbb{R}^{d_{m'}} \rightarrow \mathbb{R}^{d_\mathcal{D}^{(i)}} \) is applied to match the dimension, where $d_\mathcal{D}^{(i)}$ is the input channel dimension of the $i$-th layer's weight parameters $\mathcal{D}^{(i)}$. 
Normalization is applied to ensure consistent scaling of the mapped message. 
The modulated weights in the $i$-th layer are defined as:
\begin{equation}
\mathcal{D}^{(i)}_{mod} = \mathcal{D}^{(i)} \circ  \left(\tfrac{\mathcal{FC}^{(i)}(\mathcal{M}(m)) - \mu^{(i)}}{\sigma^{(i)} + \epsilon} \cdot \gamma^{(i)} + \beta^{(i)} \right),
\label{input:weight modulation}
\end{equation}
where $\gamma^{(i)}$ and $\beta^{(i)}$ are the $i$-th learnable parameters, while $\mu^{(i)}$ and $\sigma^{(i)}$ represent the mean and standard deviation of the message $\mathcal{FC}^{(i)}(\mathcal{M}(m))$. 

\begin{figure}[t!]
    \centering
    \includegraphics[width=0.45\textwidth]{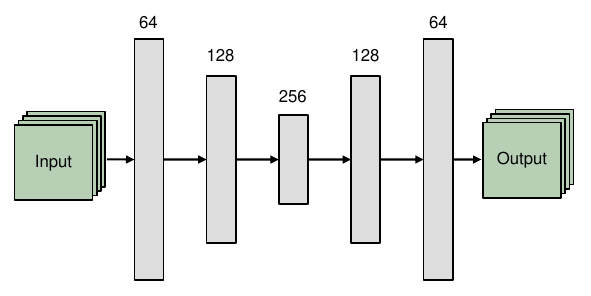}
    \vspace{-1em}
    \caption{Architecture of H.264 network. Each layer comprises a 3D convolution layer followed by a ReLU activation function. The H.264 network takes videos as input, progressively increasing the channel size to 256 during the encoding process, and then decodes the features with 3D transposed convolution layers to output results mimicking the H.264 compression.}
    \label{fig:h264 network architecture}
\end{figure}

\subsection{Distortion Layer}
\label{subsec:Distortion Layer}

Robustness against various video attacks is essential yet remains challenging for existing watermarking methods~\cite{luo2023dvmark, masoumi2013blind}. 
To address this, we apply random distortions to the generated video in every iteration.
We employ two types of distortions: spatial distortions and compression distortions. 
Spatial distortions refer to pixel-level distortions within each frame, such as cropping and blurring, which can be easily integrated.
However, the non-differentiable nature of H.264 compression complicates training. To overcome this, we design an H.264 network inspired by DVMark~\cite{luo2023dvmark} to mimic real compression, improving robustness.
Figure~\ref{fig:h264 network architecture} illustrates the architecture of our H.264 network. Using a simple structure, the network is trained to generate H.264-compressed videos. Once trained, the H.264 network is frozen and integrated into the distortion layer. Since the H.264 network is differentiable, the diffusion model is trained to improve robustness against the H.264 compression.

\subsection{Training Objectives}
\label{subsec:Training Objectives}

We design our objective function to balance the trade-off between visual quality and watermark decoding accuracy. 
We employ binary cross-entropy loss between the random message \( m \) and the extracted message \( \hat{m} \) from the generated video to ensure accurate watermark decoding:
\begin{multline}
\mathcal{L}_{msg} = -\frac{1}{N} \sum_{i=1}^{N} (m_i \cdot \log(\sigma(\hat{m}_i)) \\ + (1 - m_i) \cdot \log(1 - \sigma(\hat{m}_i))),
\label{input:watermark loss}
\end{multline}
where $N$ is the total number of message bits, while $\sigma(\cdot)$ denotes the sigmoid function. 
In addition, to enhance the visual quality of the generated video, we choose the Watson-VGG perceptual loss~\cite{czolbe2020loss}, which is designed to avoid blurring effects and visual artifacts of generated images:
\begin{equation}
\mathcal{L}_{vgg} = \text{VGG}(\mathcal{D}_{mod}({\mathcal{M}}(m),{\mathcal{E}}(v)), v),
\label{input:perceptual loss}
\end{equation}
where $\mathcal{D}_{mod}$, $\mathcal{M}$, and $\mathcal{E}$ denote the modulated latent decoder, mapping network, and latent encoder, respectively, while $m$ and $v$ are the random message and input video. 

Furthermore, we propose a weighted patch loss to address localized artifacts in the generated video. 
We first compute the Mean Absolute Error (MAE) for video patches, resulting in a loss score map normalized between 0 and 1 using the softmax function. 
By multiplying the patch MAE by the score map, we create a weighted patch loss that places more emphasis on areas with higher MAE, where artifacts are likely to occur.
The weighted patch loss is as follows:
\begin{equation}
\begin{aligned}
\mathcal{L}_{patch} &= \frac{1}{P} \sum_{i=1}^{P} |\hat{v}_i - v_i| \cdot  \text{softmax}(|\hat{v}_i - v_i|),
\end{aligned}
\label{input:weighted patch loss}
\end{equation}
where $P$ denotes the number of patches. 
The final objective can be formulated as:
\begin{equation}
\mathcal{L} = \lambda_{msg}\mathcal{L}_{msg}
+ \lambda_{vgg}\mathcal{L}_{vgg} + \lambda_{patch}\mathcal{L}_{patch},
\label{input:full loss}
\end{equation}
where $\lambda_{msg}$, $\lambda_{vgg}$, and $\lambda_{patch}$ are hyper-parameters that balance three loss functions.

\begin{table*}[t!]
\centering
\caption{Quantitative results for Open-Sora~\cite{opensora} and DynamiCrafter~\cite{xing2025dynamicrafter}}
\setlength{\tabcolsep}{1.7pt} 
\begin{tabular}{lcccccccccccccc}
\toprule
\multicolumn{2}{c}{} & \multicolumn{6}{c}{Open Sora} & \multicolumn{1}{c}{} & \multicolumn{6}{c}{DynamiCrafter} \\ 
\cmidrule{3-8} \cmidrule{10-15}
Method & Capacity (Bit) & Bits Acc.(\%) $\uparrow$ & PSNR $\uparrow$ & SSIM $\uparrow$ & LPIPS $\downarrow$ & tLP $\downarrow$ & FVD $\downarrow$ & & Bits Acc.(\%) $\uparrow$ & PSNR $\uparrow$ & SSIM $\uparrow$ & LPIPS $\downarrow$ & tLP $\downarrow$ & FVD $\downarrow$ \\ 
\midrule
HiDDeN~\cite{zhu2018hidden} & \multirow{1}{*}{32} & {99.78} & {30.19} & {0.877} & {0.237} & {5.310} & {2980.7} & & 99.82 & {30.83} & {0.888} & {0.178} & {7.596} & {1947.7} \\ 
\midrule
WAM~\cite{sander2024watermark} & \multirow{1}{*}{32} & \textbf{99.99} & {30.84} & {0.917} & {0.170} & {2.738} & {435.4} & & \textbf{99.99} & {30.92} & {0.911} & {0.156} & {3.777} & {317.1}  \\ 
\midrule
\multirow{2}{*}{Blind~\cite{masoumi2013blind}} & \multirow{1}{*}{32} & {99.71} & {29.88} & {0.888} & {0.181} & {2.615} & {363.2} & & {99.63} & {29.45} & {0.882} & {0.179} & {2.348} & {389.5} \\ 
 & \multirow{1}{*}{96} & {99.12} & {28.32} & {0.842} & {0.194} & {2.937} & {401.2} &  & {99.05} & {28.05} & {0.854} & {0.187} & {3.019} & {416.3} \\ 
\midrule
\multirow{2}{*}{REVMark~\cite{zhang2023novel}} & \multirow{1}{*}{32} & \textbf{99.99} & {29.26} & {0.905} & {0.128} & {0.317} & {133.6}  & & \textbf{99.99} & {29.02} & {0.913} & {0.113} & {0.548} & {130.9}  \\ 
& \multirow{1}{*}{96} & \underline{99.98} & {29.24} & {0.891} & {0.134} & {0.497} & {249.1}  & & \textbf{99.99} & {28.85} & {0.898} & {0.114} & {0.631} & {189.1} \\ 
\midrule
\multirow{2}{*}{Stable Signature~\cite{fernandez2023stable}} & \multirow{1}{*}{32} & {99.96} & {27.33} & {0.821} & {0.211} & {3.387} & {1094.8}  & & {99.02} & {26.78} & {0.797} & {0.223} & {4.353} & {954.8} \\ 
 & \multirow{1}{*}{96} & {95.88} & {26.46} & {0.803} & {0.259} & {4.019} & {1688.5} &  & {95.68} & {26.32} & {0.788} & {0.240} & {4.603} & {1262.8} \\ 
\midrule
\multirow{2}{*}{WOUAF~\cite{kim2024wouaf}} & \multirow{1}{*}{32} & {99.28} & {28.62} & {0.873} & {0.166} & {4.856} & {993.7} &  & {99.62} & {28.38} & {0.881} & {0.125} & {5.763} & {1469.4} \\ 
& \multirow{1}{*}{96} & {96.25} & {28.13} & {0.858} & {0.177} & {5.053} & {1196.2} &  & {96.22} & {28.23} & {0.877} & {0.130} & {5.989} & {1523.2} \\
\midrule
VideoShield~\cite{hu2025videoshield} & \multirow{1}{*}{512} & {99.52} & - & - & - & - & - &  & {99.40} & - & - & - & - & - \\ 
\midrule
VideoMark~\cite{hu2025videomark} & \multirow{1}{*}{512} & {99.45} & - & - & - & - & - &  & {99.13} & - & - & - & - & - \\ 
\midrule
\midrule
    \multirow{3}{*}{\textbf{LVMark}~\textbf{(Ours)}} & \multirow{1}{*}{32} & {\textbf{99.99}} & {\textbf{32.65}} & {\textbf{0.938}} & {\textbf{0.094}} & {\textbf{0.167}} & {\textbf{63.70}} &  & {\textbf{99.99}} & {\textbf{32.22}} & {\textbf{0.932}} & {\textbf{0.058}} & {\textbf{0.141}} & {\textbf{43.27}}  \\ 
 & \multirow{1}{*}{96} & {99.96} & {\underline{32.60}} & {\underline{0.935}} & {\underline{0.099}} & {\underline{0.198}} & {\underline{70.46}} &  & {\textbf{99.99}} & {\underline{32.06}} & {\underline{0.925}} & {\underline{0.063}} & {\underline{0.169}} & {\underline{46.83}}  \\ 
 & \multirow{1}{*}{512} & {{99.81}} & {31.74} & {0.918} & {0.105} & {0.229} & {83.98} & & {\underline{99.98}} & {31.63} & {0.914} & {0.066} & {0.189} & {68.94} \\ 
\bottomrule
\end{tabular}
\label{tab:Quantitative Results}
\end{table*}

\begin{figure*}[t!]
    \begin{center}
        \includegraphics[width=1.0\textwidth ]{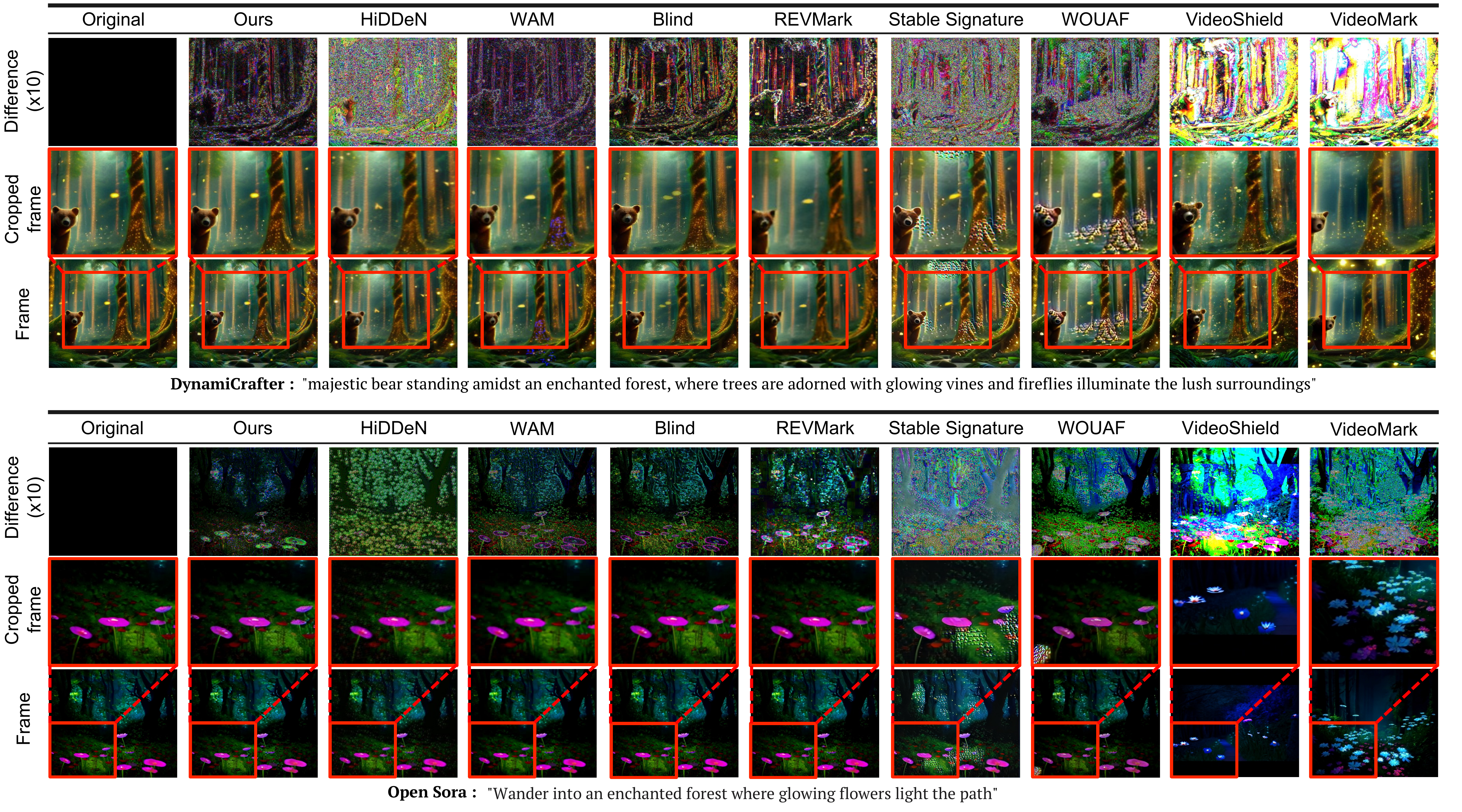}
    \end{center}
    \vspace{-1.5em}
    \caption{Qualitative results. We show the visual quality of generated videos with baseline watermarking methods. The first row shows the crop of each image, while the second and third rows show the video frame itself and the difference map ($\times$10) between the original and each method. Note that, unlike other approaches, VideoShield~\cite{hu2025videoshield} and VideoMark~\cite{hu2025videomark} embed watermarks in the latent noise of the diffusion model, generating a video different from the original.}
    \label{fig:qualitative_2}
\end{figure*}

\section{Experiments}

\subsection{Dataset}
\label{subsec:Dataset}
\noindent \textbf{Training Dataset.} We train LVMark and our H.264 network on Panda-70M dataset~\cite{chen2024panda}, a large-scale collection of video frames recognized for its diversity across various content types. 
We randomly select 10K videos from the dataset and sample 8 frames from each video, resizing each frame to a resolution of $256 \times 256$ for training.
To train the H.264 network, we obtain the ground truth video by compressing videos with H.264 using a Constant Rate Factor (CRF) of 23.

\vspace{0.5em}
\noindent \textbf{Prompt Dataset.} We evaluate text-to-video generation on the Vidprom dataset~\cite{wang2024vidprom}, using prompts generated by GPT-4~\cite{achiam2023gpt}. 
To evaluate our method, we randomly sample 100 prompts from the dataset, which do not contain personally identifiable information or offensive content.

\subsection{Implementation Details}
\label{subsec:Implementation Details}

Video diffusion models are generally classified into two categories: DiT-based~\cite{peebles2023scalable} and U-Net-based~\cite{ronneberger2015u} architectures.
To evaluate our method, we utilize Open-Sora~\cite{opensora} for the DiT architecture and DynamiCrafter~\cite{xing2025dynamicrafter} for the U-Net architecture, both of which are state-of-the-art open-source models, with Open-Sora excelling in text-to-video generation and DynamiCrafter in text-and-image-to-video generation.
Our method is trained for over 40K iterations using the AdamW optimizer~\cite{loshchilov2017decoupled} on a single A100 GPU.
We modulate 50\% of the latent decoder layers to embed the message. For fair comparisons, we apply Bose–Chaudhuri–Hocquenghem (BCH) codes, a type of error correction code, to all baseline watermarking methods under the same setting to detect and correct minor bit errors. The generated videos have a resolution of \(512 \times 320\) for DynamiCrafter and \(640 \times 640\) for Open-Sora, with both models generating 16 frames. For optimization, we set $\lambda_{msg} = 0.8$, $\lambda_{vgg} = 0.7$, $\lambda_{patch} = 10$ in Eq.~\ref{input:full loss}. Furthermore, we set patch size to 8 and employ 100 random messages.
Additionally, since H.264 compression is non-differentiable, we pre-train a H.264 network to approximate the behavior of the actual H.264 compression. Our H.264 network is trained to generate H.264-compressed videos by minimizing the Mean Squared Error and LPIPS loss~\cite{zhang2018unreasonable} between the network's outputs and the actual H.264-compressed videos.

\subsection{Evaluation}
We evaluate three essential aspects of digital watermarking, which exhibit trade-offs: 1) \textit{invisibility}, which is investigated by PSNR, SSIM, and LPIPS~\cite{zhang2018unreasonable} to measure the visual quality of each frame. To evaluate the video quality, we report tLP~\cite{chu2020learning} and Fr$\acute{\text{e}}$chet Video Distance (FVD)~\cite{unterthiner2018towards} to measure the temporal consistency of the watermarked videos.
2) \textit{capacity}, which we investigate with 32-bit, 96-bit, and 512-bit messages. 
3) \textit{robustness}, which is categorized into image, video, and model attack. 
For image attacks, we evaluate Resize (50\%), Text Overlay (30px), Crop (50\%), Rotation ($\pi / 6$), Gaussian Blur (standard deviation = 1.0), Gaussian Noise (standard deviation = 0.05), and JPEG Compression (50\% of the original).
For video attacks, we evaluate Frame Insert (3 frames), Frame Pad (20px), Frame Swap (change 50\% positions of frames), Frame Average (average of 6 frames), Frame Drop (Drop 50\% frames), H.264 (CRF=24), and the combination of rotation, frame drop, and H.264.
For model attacks, we evaluate the effects of Gaussian noise injection (\(\sigma=0.05\), applied with 20\% probability), pruning (with 10\% probability), and retraining with a randomly assigned message.

\vspace{0.5em}
\noindent \textbf{Baselines.} For a fair comparison, we evaluate seven methods across three types of watermarking methods:
1) HiDDeN~\cite{zhu2018hidden} and WAM~\cite{sander2024watermark} are watermarking methods for images.
2) Blind~\cite{masoumi2013blind} and REVMark~\cite{zhang2023novel} are watermarking methods for videos. 
3) Stable Signature~\cite{fernandez2023stable} and WOUAF~\cite{kim2024wouaf} for are watermarking methods for image diffusion models.
4) VideoShield~\cite{hu2025videoshield} and VideoMark~\cite{hu2025videomark} are watermarking methods for video diffusion models.
For both Stable Signature and WOUAF, we fine-tune Open-Sora~\cite{opensora} and DynamiCrafter~\cite{xing2025dynamicrafter} using the same training set as our method. 
Additionally, we decode the message from each video frame for image-level watermarking methods~\cite{zhu2018hidden, sander2024watermark, fernandez2023stable, kim2024wouaf}.

\subsection{Video Quality and Bit Accuracy}

There is a trade-off between video quality and bit accuracy, making it a key challenge in digital watermarking to improve both simultaneously. 
Consequently, for a fair comparison, we measure the video quality when bit accuracy is maintained close to 100\%.
As shown in Table~\ref{tab:Quantitative Results}, we compare the generated video quality and bit accuracy of our method against other methods. 
Note that since VideoShield~\cite{hu2025videoshield} generates a different video from the original, the visual quality metrics in the table are not meaningful.
Our method outperforms other methods, especially video quality and temporal consistency. Figure~\ref{fig:qualitative_2} and Figure~\ref{fig:further_qualitative} also show our method achieves higher visual quality compared to other methods.  
We observe that image-based watermarking methods~\cite{zhu2018hidden, sander2024watermark, fernandez2023stable, kim2024wouaf} do not fully address temporal consistency, as they independently embed the invisible message into each frame of the generated video. 
The performance drop of image diffusion model watermarking methods~\cite{fernandez2023stable, kim2024wouaf} is particularly pronounced when applied to video diffusion models, as they are designed for image-based models that ignore temporal information.
Video watermarking methods~\cite{masoumi2013blind, zhang2023novel} maintain relatively high bit accuracy and video quality but exhibit inferior performance compared to our method. Although VideoShield~\cite{hu2025videoshield} and VideoMark~\cite{hu2025videomark} achieve high bit accuracy, they produce videos that are different from the original.
In contrast, our method preserves the original video, while achieving the highest bit accuracy and temporal consistency.

Furthermore, we conduct a user study with 100 participants recruited via Amazon Mechanical Turk (AMT)~\cite{amazon_mturk}. Informed consent is obtained from all participants through the AMT platform prior to their participation. All data are collected anonymously, and no personally identifiable information is gathered. Participants compare videos from our method and all baselines across six randomly selected scenes and select the one most similar to the original. As shown in Table~\ref{tab:user_study}, our method receives the highest average preference score, confirming its superior video quality. 

\begin{table}[t!]
\centering
\setlength{\tabcolsep}{1.5em} 
\caption{The result of User Study}
\label{tab:user_study}
\begin{tabular}{@{}lcc@{}}
\toprule
\multicolumn{1}{l}{Methods} & OpenSora(\%) & DynamiCrafter(\%) \\
\midrule
HiDDeN~\cite{zhu2018hidden}         & 5.67  & 5.33 \\
WAM~\cite{sander2024watermark}         & 6.33  & 6.67 \\
Blind~\cite{masoumi2013blind}        & 2.00  & 2.67  \\
REVMark~\cite{zhang2023novel}         & 4.67  & 3.33  \\
Stable Signature~\cite{fernandez2023stable}         & 3.33 & 2.33  \\
WOUAF~\cite{kim2024wouaf}         & 1.33 & 1.33  \\
VideoShield~\cite{hu2025videoshield}         & 0.33  & 0.33  \\
VideoMark\cite{hu2025videomark}   & 0.33  & 1.33  \\
\textbf{LVMark (Ours)}  & \textbf{76.00} & \textbf{76.67} \\
\bottomrule
\end{tabular}
\end{table}

\begin{figure*}[t!]
    \begin{center}
        \includegraphics[width=1.0\textwidth ]{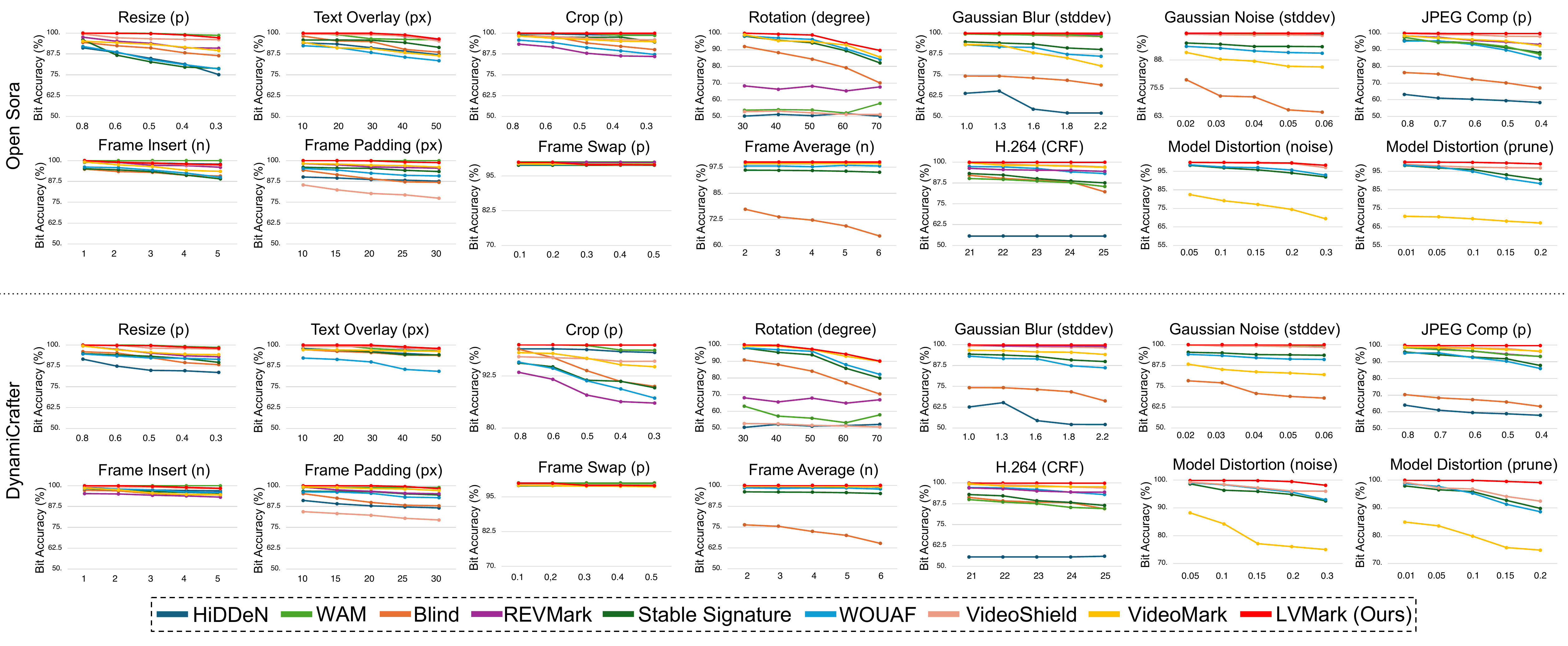}
    \end{center}
    \vspace{-1.5em}
    \caption{Visualization of robustness for various attacks. This figure shows the robustness under various attacks with varying intensities. The top and bottom respectively correspond to experiments for 32-bit. We use decoded bit accuracy to measure robustness. The (p) in each graph title represents the probability, indicating the proportion of frames subjected to the specified attack. (stddev) denotes the standard deviation of the Gaussian blur and noise applied during the attack, while (n) refers to the number of frames affected by the attack. (px) means pixel.}
    \label{fig:robustness_32}
\end{figure*}

\subsection{Capacity of Watermark} 
In this section, we analyze the video quality and bit accuracy based on the length of the embedded message. 
A higher watermark bit capacity enables the registration and tracking of more users’ copyright information, enhancing the practicality of watermarking methods.
We compare our method with others using 100 different messages for 32-bit and 96-bit capacities. Furthermore, we embed 100 different 512-bit messages into our method, demonstrating its ability to maintain both video quality and bit accuracy at higher capacities.
As shown in Table~\ref{tab:Quantitative Results}, all methods exhibit a decline in bit accuracy and visual quality as the message length increases. 
Notably, our method achieves superior video quality, even when embedding a 512-bit message, outperforming other methods with a 32-bit capacity.
Therefore, experiments with higher capacities for other methods are unnecessary.
This highlights the superiority of our method in embedding high-capacity watermarks while preserving video quality better than other approaches.

\begin{table}[t!]
\vspace{-0.5em}
\centering
\setlength{\tabcolsep}{0.55em} 
\caption{Evaluation of different video resolutions and lengths}
\label{tab:video length and size}
\begin{tabular}{lccc}
\toprule
& \multicolumn{3}{c}{Bit Accuracy (\%) $\uparrow$} \\ 
\midrule
Length(T) \( \backslash \) Resolution(H $\times$ W) & 144 $\times$ 256 & 240 $\times$ 426 & 640 $\times$ 640 \\ 
\midrule
\multicolumn{1}{c}{T = 16}  & 99.78 & 99.80 & 99.81  \\ 
\multicolumn{1}{c}{T = 32}  & 99.80 & 99.81 & 99.81  \\ 
\multicolumn{1}{c}{T = 64}  & 99.81 & 99.81 & 99.82  \\ 
\multicolumn{1}{c}{T = 100}  & 99.81 & 99.82 & 99.83  \\ 
\bottomrule
\end{tabular}
\end{table}

\subsection{Performance on Larger Videos} 
Generating long, high-resolution video is a critical capability of video diffusion models~\cite{xing2023survey}. Consequently, it is essential for watermarking methods to maintain robust performance on long, high-resolution videos. Table~\ref{tab:video length and size} demonstrates that our method performs consistently well across various frame lengths. Notably, the bit accuracy slightly increases with higher resolutions, as the increased capacity of the video allows for more effective watermark embedding.

\begin{table*}[t!]
\setlength{\tabcolsep}{0.095em}
\renewcommand{\arraystretch}{1}
\centering
\caption{Robustness comparisons under image and video attacks (Best in bold, second in underline).\\
Res. denotes Resize, Over. denotes Text Overlay, Rot. denotes Rotation, F. Ins. denotes Frame Insert, F. Pad denotes Frame Pad, \\
F. Swap denotes Frame Swap, F. Avg. denotes Frame Average, F. Drop denotes Frame Drop, Comb. denotes  Combined Attack.}
\label{tab:robustness}
\begin{tabular}{llccccccccccccccccc}
\toprule
\multicolumn{1}{c}{\multirow{3}{*}{}} & \multicolumn{1}{c}{\multirow{1}{*}{}} & \multirow{2}{*}{Capacity} & \multicolumn{1}{c}{} & \multicolumn{7}{c}{Image Attack} & & \multicolumn{6}{c}{Video Attack} \\
\cmidrule{5-11} \cmidrule{13-18}
\multicolumn{1}{c}{} & Methods & (Bit) & None & \begin{tabular}[c]{@{}c@{}} Res.\\ (50\%)\end{tabular} & \begin{tabular}[c]{@{}c@{}} Over. \\ (30px)\end{tabular} & \begin{tabular}[c]{@{}c@{}}Crop\\ (50\%)\end{tabular} & \begin{tabular}[c]{c}Rot.\\ ($+ \pi / 6$)\end{tabular} & \begin{tabular}[c]{c}Blur\\ (std = 1.0)\end{tabular} & \begin{tabular}[c]{c}Noise\\ (std = 0.05)\end{tabular} &\begin{tabular}[c]{@{}c@{}}JPEG\\ (50\%)\end{tabular} & & \begin{tabular}[c]{c}F. Ins.\\ (n=3)\end{tabular} & \begin{tabular}[c]{c}F. Pad\\ (20px)\end{tabular} & \begin{tabular}[c]{c}F. Swap\\ (p=0.5)\end{tabular} & \begin{tabular}[c]{c}F. Avg.\\ (n=6)\end{tabular} & \begin{tabular}[c]{c}F. Drop\\ (p=0.5)\end{tabular} & \begin{tabular}[c]{c}H.264 \\ (CRF=24)\end{tabular} & \begin{tabular}[c]{@{}c@{}}Comb.\end{tabular}  \\ 
\midrule
\multirow{10}{*}{{\rotatebox[origin=c]{90}{Open Sora}}} 
& HiDDeN~\cite{zhu2018hidden} & 32 & {99.10} & 84.81 & 91.24 & {99.07} & {50.24} & {63.89} & 99.08 & {59.47} && 98.76 & 88.61 &  {99.10} & {98.93} & {99.75} & {55.66} & {55.03} \\ 
& WAM~\cite{sander2024watermark} & 32 & \textbf{99.99} & \underline{99.66} & 96.58 & \underline{99.94} & {53.74} & \underline{99.95} & \underline{99.94} & {91.94} && \textbf{99.99} & \underline{99.84} & \textbf{99.99} & \underline{99.97} & \textbf{99.99} & {87.60} & {58.03} \\ 
& Blind~\cite{masoumi2013blind} & 96 & {99.12} & 91.12 & 94.92 & {88.51} & {87.32} & {69.24} &65.95 &{67.53} && 89.96 & 89.14 & {89.51} & {59.53} & {90.12} & {84.13} & {88.19} \\ 
& REVMark~\cite{zhang2023novel} & 96 & \underline{99.98} & 93.86 & 99.16 & {82.91} & {59.47} & {99.59} & 99.93& {94.65} && 95.86 & 95.68 & {99.74} & \textbf{99.98} & {98.09} & {93.49} & {81.20} \\ 
& Stable Signature~\cite{fernandez2023stable} & 96  & {95.88} & 75.72 & 95.81 & {92.49} & {94.15} & {89.80} & 94.08&{83.61} && 91.53 & 94.05 & {95.88} & {93.29} & {94.14} & {82.63} & {76.32} \\ 
& WOUAF~\cite{kim2024wouaf} & 96 & {96.25} & 74.07 & 78.39 & {88.51} & {95.33} & {87.80} & 91.41& {87.75} && 92.63 & 90.25 & {96.25} & {94.56} & {93.26} & {90.15} & {86.32} \\ 
& VideoShield~\cite{hu2025videoshield} & 512 & {99.52}  & 96.59 & 99.07 & {96.93} & {52.89} & {99.49} & 99.13 & {98.12} && 98.99 & 73.44 & {99.34} & {99.64} & {93.16} & {97.92} & {75.25} \\ 
& VideoMark~\cite{hu2025videomark} & 512 & 99.45 & 93.34 & 88.71 & 96.10 & 99.27  & 93.19 & 83.26 & 93.56  && 96.31 & 96.71 & 99.16 & 98.94 & 95.87 & 96.47 & 86.86 \\
\cmidrule{2-19}
& \multirow{3}{*}{\textbf{LVMark}~\textbf{(Ours)}} & 32 & {\textbf{99.99}} &\textbf{99.82} & \textbf{99.93} & {\textbf{99.96}} & {\textbf{99.95}} & {\textbf{99.96}} & \textbf{99.95} &{\textbf{99.84}} && \underline{99.93} & \textbf{99.90} & {\textbf{99.99}} & {\textbf{99.98}} & {\textbf{99.99}} & {\textbf{99.84}} & {\textbf{99.83}}  \\
& & 96 & {99.96} & 99.54 & \underline{99.48} & {99.91} & {\underline{99.82}} & {\underline{99.95}} & 99.19 & {\underline{99.77}} && 98.41 & 99.60 & {\underline{99.96}} & {99.96} & {\underline{99.96}} & {\underline{99.76}} & {\underline{99.81}} \\
& & 512 & {{99.81}} & 99.33 & 99.04 & {{99.74}} & {99.51} & {99.66} & 99.66 & {99.19} && 98.36 & 99.28 & {99.81} & {{99.04}} & {{99.24}} & {98.20} & {99.19}  \\
\midrule
\midrule
\multirow{10}{*}{{\rotatebox[origin=c]{90}{DynamiCrafter}}} 
& HiDDeN~\cite{zhu2018hidden} & 32 & {99.02} & 86.12 & 93.42 & {98.82} & {50.38} & {62.64} & 99.80 & {58.81} && 97.37 & 87.94 & {99.02} & {98.91} & \underline{99.81} & {56.01} & {55.82} \\ 
& WAM~\cite{sander2024watermark} & 32 & \textbf{99.99} & \underline{99.54} & \underline{99.76} & {99.78} & {63.11} & \textbf{99.98} & \underline{99.94} & {94.34} && \textbf{99.99 }& 98.93 & \textbf{99.99} & \underline{99.98} & \textbf{99.99} & {85.27} & {64.20} \\ 
& Blind~\cite{masoumi2013blind} & 96 & {99.05} &  92.79 &  97.54 & {84.26} & {85.75} & {69.29} & 68.95 & {61.48} &&  86.37 &  88.67 & {85.19} & {57.25} & {86.24} & {85.63} & {85.02}\\ 
& REVMark~\cite{zhang2023novel} & 96 & \textbf{99.99} &  92.49 &  98.99 & {81.95} & {58.27} & \underline{99.88} & 99.83 & {95.86} &&  99.98 &  98.09 & {99.79} & \textbf{99.99} & {97.83}  & {94.52} & {80.29}  \\ 
& Stable Signature~\cite{fernandez2023stable} & 96 & {95.68} &  88.09 &  94.15 & {92.11} & {93.86} & {89.03} & 93.88 &{84.11} &&  94.14 & 91.41 & {95.68} & {93.23} & {94.01} & {81.17} & {75.32} \\ 
& WOUAF~\cite{kim2024wouaf} & 96 & {96.22} & 73.76 & 87.91 & {86.25} & {94.84} & {85.80} &  91.33 & {87.47} && 95.13 & 94.97 & {96.22} & {94.42} & {93.10} & {88.90} & {85.09} \\ 
& VideoShield~\cite{hu2025videoshield} & 512 & {99.40} & 98.94 & 99.49 & {96.65} & {50.11} & {99.20} & 98.78 & {97.99} && 97.67 & 60.53 & {99.26} & {99.35} & {92.57} & {97.60} & {71.96}   \\ 
& VideoMark~\cite{hu2025videomark} & 512 & 99.13 & 93.29 & 93.33 & 96.51 & 99.13 & 93.33 & 76.49 & 93.54 && 98.48 & 98.93 & 99.54 & 98.62 & 94.74  & 96.60  &  88.71  \\ 
\cmidrule{2-19}
& \multirow{3}{*}{\textbf{LVMark}~\textbf{(Ours)}} & 32 & {\textbf{99.99}} & \textbf{99.59} & \textbf{99.98} & {\textbf{99.98}} & {\textbf{99.98}} & {\textbf{99.98}} & \textbf{99.95} & {\textbf{99.98}} && \underline{99.98} & \textbf{99.99} & {\textbf{99.99}} & {\textbf{99.99}} & {\textbf{99.99}} & {\textbf{99.98}} & {\textbf{99.74}}  \\
& & 96 & \textbf{99.99} & 99.46 & \textbf{99.98} & {\textbf{99.98}} & {\textbf{99.98}} & {\textbf{99.98}} & 99.74 & {\underline{99.95}} && 99.23 & \underline{99.98} & {\textbf{99.99}} & {{99.97}} & {\textbf{99.99}} & {\underline{99.94}} & {\underline{99.50}}   \\
        & & 512 & {\underline{99.98}} & 99.34 & 99.56 & \underline{99.90} & \underline{99.93} & {99.74} & 99.36 &{99.94} && 98.46 & 99.44 & {\underline{99.98}} & {{99.04}} & {{99.36}} & {99.75}  & {99.16}  \\
\bottomrule
\end{tabular}
\end{table*}

\begin{table*}[htb!]
\renewcommand{\arraystretch}{1.}
\centering
\caption{Robustness comparisons under model attacks (Best in bold, second in underline).}
\label{tab:robustness_model}
\begin{tabular}{llccccc}
\toprule
\multicolumn{1}{c}{\multirow{3}{*}{}} & \multicolumn{1}{c}{\multirow{1}{*}{}} & \multirow{2}{*}{Capacity} & \multicolumn{1}{c}{} & \multicolumn{3}{c}{Model Attack} \\
\cmidrule{5-7}
\multicolumn{1}{c}{} & Methods & (Bit) & None & \begin{tabular}[c]{c}Gaussian / (PSNR) \\ (p=0.2)\end{tabular}  & \begin{tabular}[c]{c}Prune / (PSNR)\\ (p=0.1)\end{tabular} & \begin{tabular}[c]{@{}c@{}}Retrain / (PSNR)\end{tabular} \\ 
\midrule
\multirow{6}{*}{{\rotatebox[origin=c]{90}{Open Sora}}} 
& Stable Signature~\cite{fernandez2023stable} & 96  & {95.88} & {93.15 / (24.67)} & {93.99 / (20.03)} & {50.19 / (26.51)}\\ 
& WOUAF~\cite{kim2024wouaf} & 96 & {96.25} &  {93.68 / (26.45)} & {92.93 / (19.67)} & {85.47 / (28.15)}\\ 
& VideoShield~\cite{hu2025videoshield} & 512 & {99.52} & {96.32} / \;\;\; - \;\;\; & {97.19} / \;\;\; - \;\;\; & 98.93 / \;\;\; - \;\;\;  \\ 
& VideoMark~\cite{hu2025videomark} & 512 & {98.85} & {56.92} / \;\;\; - \;\;\; & {55.35}  / \;\;\; - \;\;\; & 64.43 / \;\;\; - \;\;\; \\ \cmidrule{2-7}
& \multirow{3}{*}{\textbf{LVMark}~\textbf{(Ours)}} & 32 & {\textbf{99.99}} & {\textbf{99.52 / (29.02)}} & {\textbf{99.85 / (22.72)}} & {\textbf{99.16 / (32.69)}}  \\
& & 96 & {99.96} & {\underline{99.46 / (28.97)}} & {\underline{99.84 / (22.70)}} & {\underline{98.99 / (32.58)}} \\
& & 512 & {{99.81}} & {99.29 / (28.88)} & {99.81 / (22.53)} & {97.67 / (31.72)}  \\
\midrule
\multirow{6}{*}{{\rotatebox[origin=c]{90}{DynamiCrafter}}} 
& Stable Signature~\cite{fernandez2023stable} & 96 & {95.68} & {93.68 / (23.92)} & {94.15 / (19.94)} & {50.24 / (26.35)}\\ 
& WOUAF~\cite{kim2024wouaf} & 96 & {96.22} &  {94.13 / (25.11)} & {94.07 / (18.96)}  & {83.43 / (28.21)} \\ 
& VideoShield~\cite{hu2025videoshield} & 512 & {99.40} & {96.05} / \;\;\; - \;\;\; & {96.81} / \;\;\; - \;\;\; & 97.19 / \;\;\; - \;\;\;  \\ 
& VideoMark~\cite{hu2025videomark} & 512 & {99.13} & {61.21} / \;\;\; - \;\;\; & {52.94}  / \;\;\; - \;\;\; & 59.73 / \;\;\; - \;\;\; \\
\cmidrule{2-7}
& \multirow{3}{*}{\textbf{LVMark}~\textbf{(Ours)}} & 32 & {\textbf{99.99}}  & {\textbf{99.74 / (26.45)}} & {\textbf{99.90 / (21.24)}} & {\textbf{97.36 / (32.25)}}  \\
& & 96 & {99.99} & {\underline{99.72 / (26.41)}} & {\underline{99.87 / (21.19)}} & {\underline{97.24 / (32.02)}}  \\
        & & 512 & {\underline{99.98}}  & {99.62 / (26.26)} & {99.76 / (21.03)} & {95.73 / (31.59)}  \\
\bottomrule
\end{tabular}
\end{table*}

\subsection{Robustness of Watermark} 
\noindent \textbf{Image and Video Distortions.} 
A key priority in digital watermarking is robustness against distortions. 
In this section, we evaluate the robustness to distortions in videos and their individual frames.
Table~\ref{tab:robustness} shows the bit accuracy of our method against various distortions, compared to other methods. 
We note that since all methods, except Blind~\cite{masoumi2013blind} and REVMark~\cite{zhang2023novel}, are applied independently per frame, it is naturally robust against frame swap. 
By combining spatio-temporal information from the 3D wavelet domain and the RGB domain, our watermark decoder achieves high bit accuracy across all evaluated attacks.
It is particularly effective against temporal manipulations such as frame swapping and frame dropping.
Additionally, our method shows exceptional robustness against compressions such as JPEG and H.264.
Compression methods tend to remove high-frequency components, which leads to in the loss of embedded information.
In contrast, our method leverages low-frequency components along with RGB video features, making it inherently more robust to the compression. Furthermore, as shown in Figure~\ref{fig:robustness_32}, our method consistently outperforms other methods on both Open-Sora and DynamiCrafter, particularly under stronger and more challenging attack conditions.

\begin{figure}[t!]
    \begin{center}
        \includegraphics[width=0.95\linewidth]{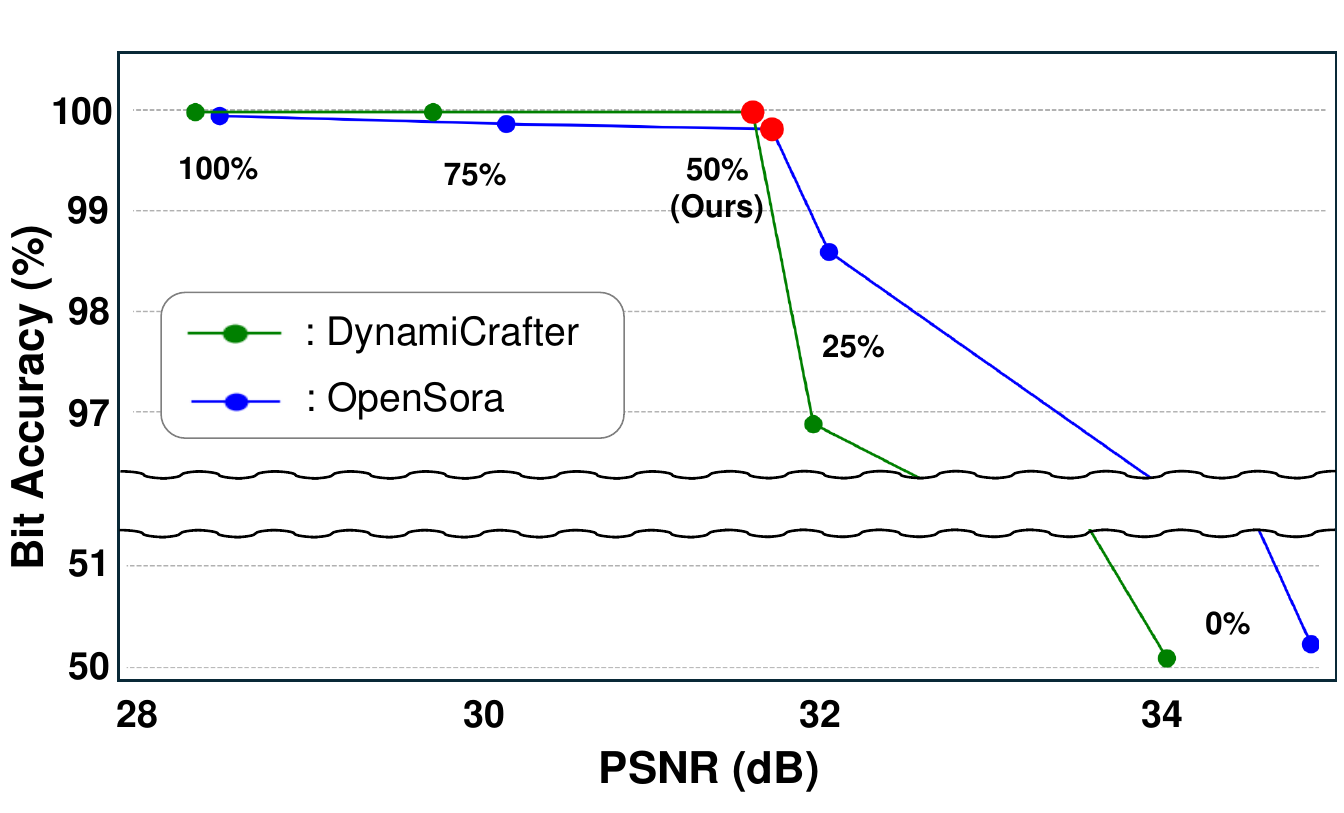}
    \end{center}
    \vspace{-2em}
    \caption{The impact of weight modulation rate. Each point represents the metrics for different modulation rates: 0\%, 25\%, 50\%, 75\%, and 100\%. We evaluate the modulation rate that achieves the best performance in terms of both visual quality and bit accuracy, using PSNR and bit accuracy.}
    \label{fig:modulation}
\end{figure}

\vspace{0.5em}
\noindent \textbf{Model Distortions.}
In the scenario where an unauthorized user illicitly utilizes the diffusion model, they manipulate the model to generate videos that bypass watermark detection. 
Even if significant distortions, such as Gaussian noise injection or layer pruning, lead to degradation in video quality, the watermark should still be decoded under such attacks.
To address this, we evaluate the robustness against various model distortions.
Additionally, unauthorized users attempt to retrain the model using their own watermark key to override the original watermark.
To simulate this, we also evaluate bit accuracy when the model is retrained with a different watermark key.
Table~\ref{tab:robustness_model} shows bit accuracy and PSNR under model attacks. We note that HiDDeN~\cite{zhu2018hidden}, WAM~\cite{sander2024watermark}, Blind~\cite{masoumi2013blind}, and REVMark~\cite{zhang2023novel} are not applicable to model attacks since they are post-processing watermarking methods. VideoShield~\cite{hu2025videoshield} and VideoMark~\cite{hu2025videomark} do not preserve the original videos, leaving the PSNR unreported. 
Our method ensures the reliable extraction of messages even when the model is distorted. Model distortions inherently alter parameters, leading to variations and degradation in the spatial and temporal details of the generated videos.
The superior bit accuracy of our method under identical attack conditions is primarily attributed to the proposed Importance-based Weight Modulation, which effectively mitigates visual degradation. In contrast to other methods that perturb all model weights, our approach selectively fine-tunes specific layers that have minimal impact on the visual generation process. This ensures parameter consistency with the original model, resulting in high visual quality. Preserving the visual and structural integrity of the generated videos allows the embedded watermark to survive severe model distortions.
Moreover, our decoder plays a pivotal role in achieving this robustness. By extracting watermark features from both the wavelet and RGB domains, and by training with diverse watermark keys and simulated layer distortions, the decoder becomes highly resilient to unpredictable input variations. This strategy ensures reliable watermark recovery even when the model is severely distorted.
Therefore, under model distortions including Gaussian noise, pruning, and retraining, our method achieves higher bit accuracy compared to other methods.

\begin{figure}[t!]
    \vspace{0.4em}
    \centering
    \includegraphics[width=0.45\textwidth]{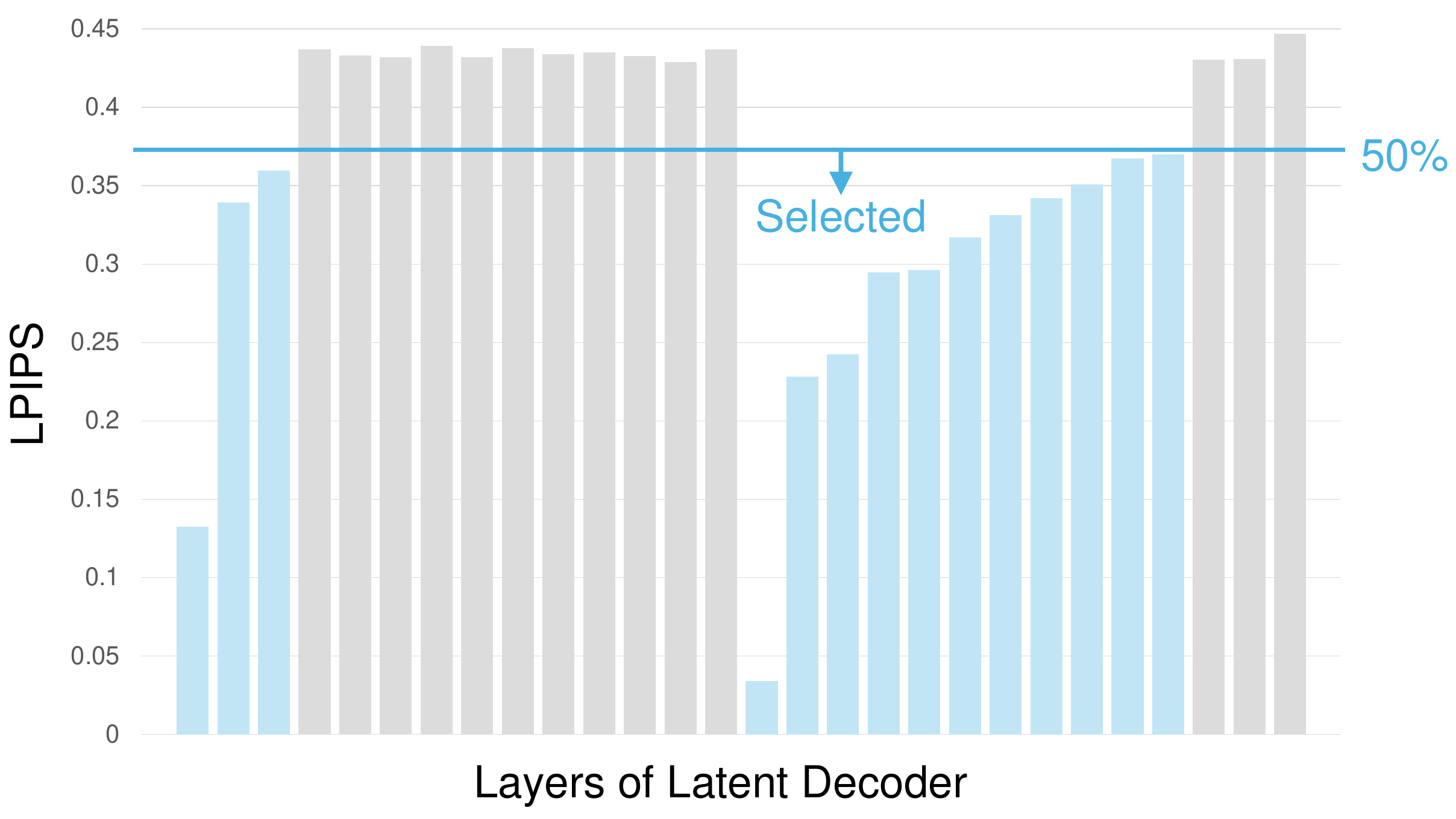}
    \vspace{-0.5em}
    \caption{Visualization of the impact of each layer on video quality. This figure shows the LPIPS performance of videos generated by adding Gaussian noise with a standard deviation of 0.2 to each layer of DynamiCrafter's latent decoder. We modulate only 50\% of layers with the lowest LPIPS values.}
    \label{fig:weight_importance}
\end{figure}

\subsection{Ablation Study}
\label{subsec:Ablation study}
For the ablation study, we follow the same setup as in Section~\ref{subsec:Implementation Details} with 512-bit messages.

\vspace{0.5em}
\noindent \textbf{Importance-based Weight Modulation.}
To ensure high video quality, we propose an importance-based weight modulation strategy to embed random watermark messages into diffusion models. Empirical analysis shows that naively modulating all layers degrades video quality. In this section, we provide additional details and experiments demonstrating its effectiveness.
As mentioned in Section~\ref{subsec:Importance-based Weight Modulation}, we selectively modulate the bottom \(k\) \% of the latent decoder layers with the least impact on video quality to embed messages while minimizing visual degradation. As shown in Figure~\ref{fig:modulation}, as the modulation rate \(k\) approaches 100\%, bit accuracy improves at the cost of video quality, whereas setting \(k\) as 0\% results in failed bit extraction, illustrating a trade-off. We find an optimal balance between bit accuracy and video quality when $k$ is set to 50.
Figure~\ref{fig:weight_importance} presents LPIPS scores for videos generated by adding Gaussian noise to individual layers of DynamiCrafter’s latent decoder. The result indicates that certain layers have minimal impact on video quality. Based on this, we selectively modulate the 50\% of layers that contribute least to degradation.
Table~\ref{tab:Weight_modulation} shows the bit accuracy and visual quality when modulating the top 50\% layers and a randomly selected 50\% layers.
The result shows that modulating either highly important or randomly chosen layers leads to degradation in both video quality and bit accuracy.
In contrast, modulating the bottom 50\% of less important layers achieves superior performance, highlighting its effectiveness for watermark embedding.

\begin{table*}[t!]
\centering
\setlength{\tabcolsep}{8pt} 
\renewcommand{\arraystretch}{1}
\caption{Experiments on importance-based weight modulation}
\label{tab:Weight_modulation}
\begin{tabular}{cccccccccc}
\toprule
\multicolumn{1}{c}{} & \multicolumn{1}{c}{} & \multicolumn{5}{c}{Video Quality Metrics} & \multicolumn{1}{c}{} & \multicolumn{2}{c}{Bit Accuracy (\%) $\uparrow$} \\
\cmidrule{3-7} \cmidrule{9-10} 
\multicolumn{1}{c}{} & \multicolumn{1}{c}{Selected Layers Importance} & PSNR $\uparrow$ & SSIM $\uparrow$ & LPIPS $\downarrow$ & tLP $\downarrow$ & FVD $\downarrow$ & & None & H.264 (CRF=24) \\ 
\midrule
\multirow{2}{*}{\begin{tabular}[c]{@{}c@{}}Open Sora\end{tabular}} 
& \multicolumn{1}{l}{Top 50\%} & 28.88 & 0.898 & 0.137 & 0.304 & 108.6 & & 99.74 & 98.12\\
& \multicolumn{1}{l}{Random 50\%} & 30.81 & 0.901 & 0.129 & 0.282 & 95.63 & & 99.76 & 98.13\\
\cmidrule{2-10}
& \multicolumn{1}{l}{Bottom 50\% \textbf{(Ours)}} & \textbf{31.74} & \textbf{0.918} & \textbf{0.105} & \textbf{0.229} & \textbf{83.98} & & \textbf{99.81} & \textbf{98.20} \\
\midrule
\multirow{2}{*}{\begin{tabular}[c]{@{}c@{}}DynamiCrafter\end{tabular}} 
& \multicolumn{1}{l}{Top 50\%} & 29.16 & 0.896 & 0.109 & 0.216 & 93.78 & & 99.95 & 99.68  \\
& \multicolumn{1}{l}{Random 50\%} & 30.28 & 0.897 & 0.097 & 0.206 & 91.14 & & 99.74 & 99.52  \\
\cmidrule{2-10}
& \multicolumn{1}{l}{Bottom 50\% \textbf{(Ours)}} & \textbf{31.63} & \textbf{0.914} & \textbf{0.066} & \textbf{0.189} & \textbf{68.94} & & \textbf{99.98} & \textbf{99.75}\\
\bottomrule
\end{tabular}
\vspace{-1em}
\end{table*}

\begin{table}[t!]
\centering
\setlength{\tabcolsep}{5pt} 
\renewcommand{\arraystretch}{1.2} 
\caption{Comparison of different frequency types}
\label{tab:Ablation on the DWT}
\begin{tabular}{llcccc}
\toprule
& DWT Type  & Bits Acc. $\uparrow$ & PSNR $\uparrow$ & tLP $\downarrow$ & FVD $\downarrow$  \\ 
\midrule
\multirow{4}{*}{\rotatebox[origin=c]{90}{Open Sora}}
& 3D DCT & 98.74 & 30.73 & 0.249 & 104.8 \\
& 3D FFT & 97.98 & 31.28 & 0.232 & 85.37  \\
& 2D DWT & 96.89 & 30.97 & 0.410 & 175.4  \\ 
\cmidrule{2-6}
& 3D DWT \textbf{(Ours)} & \textbf{99.81} & \textbf{31.74} & \textbf{0.229} & \textbf{83.98}    \\ 
\midrule
\multirow{4}{*}{\rotatebox[origin=c]{90}{\shortstack{DynamiCrafter}}}
& 3D DCT & 99.21 & 30.72 & 0.293 & 114.7  \\
& 3D FFT & 95.86 & 31.42 & 0.197 & 72.35  \\
& 2D DWT & 96.63  & 30.89  & 0.324 & 143.9  \\ 
\cmidrule{2-6}
& 3D DWT \textbf{(Ours)} & \textbf{99.98} & \textbf{31.63} & \textbf{0.189} & \textbf{68.94}    \\ 
\bottomrule
\end{tabular}
\end{table}

\begin{table}[t!]
\centering
\caption{Comparison on frequency subband and H.264 network.}
\label{tab:No H264}
\renewcommand{\arraystretch}{1.2}
\setlength{\tabcolsep}{1.4em} 
\begin{tabular}{ccccc}
\toprule
\multicolumn{1}{c}{} & \multicolumn{2}{c}{} & \multicolumn{2}{c}{Bit Accuracy (\%) $\uparrow$} \\
\cmidrule{4-5}
\multicolumn{1}{c}{} & \multicolumn{1}{c}{Frequency} & \multicolumn{1}{c}{H.264} & None & H.264 (CRF=24) \\ 
\midrule
\multirow{4}{*}{\rotatebox[origin=c]{90}{\shortstack{Open Sora}}}
& high & - & 95.12 & 85.87 \\
& high & $\checkmark$ & 95.46 & 89.40 \\
& low & - & \textbf{99.81} & 94.23 \\
\cmidrule{2-5}
& low & $\checkmark$  & \textbf{99.81} & \textbf{98.20} \\
\midrule
\multirow{4}{*}{\rotatebox[origin=c]{90}{\shortstack{DynamiCrafter}}}
& high & - & 94.92 & 84.74 \\
& high & $\checkmark$ & 94.78 & 88.93 \\
& low & -  & \textbf{99.98} & 95.89 \\
\cmidrule{2-5}
& low & $\checkmark$  & \textbf{99.98} & \textbf{99.75} \\
\bottomrule
\end{tabular}
\end{table}

\begin{table}[t!]
\centering
\setlength{\tabcolsep}{5pt} 
\renewcommand{\arraystretch}{1.2} 
\caption{Comparison of different domain types}
\label{tab:Comparison of different domain types}
\begin{tabular}{llccc}
\toprule
\multicolumn{2}{c}{} & \multicolumn{1}{c}{} & \multicolumn{2}{c}{Bit Accuracy (\%)~$\uparrow$} \\
\cmidrule{4-5} 
& Domain Type  & PSNR $\uparrow$ & None & H264 (CRF=24)  \\ 
\midrule
\multirow{12}{*}{\rotatebox[origin=c]{90}{\shortstack{Open Sora}}}
& RGB only & 30.87 & 94.87 & 86.99  \\
& LLL only & \textbf{33.85}  & 50.23 & 50.13 \\
& HHH only & 33.74  & 50.62 & 50.54 \\
& LLH + RGB & 30.91  & 96.46 & 92.96 \\
& LHL + RGB & 31.25  & 95.71 & 92.40 \\
& HLL + RGB & 31.88  & 94.08 & 92.57 \\
& HHL + RGB & 31.07  & 95.23 & 91.42 \\
& HLH + RGB & 30.94  & 95.87 & 90.97 \\
& LHH + RGB & 31.47  & 95.19 & 89.94 \\
& HHH + RGB & 30.50  & 95.46 & 89.40 \\ 
\cmidrule{2-5}
& LLL + RGB \textbf{(Ours)} & 31.74  & \textbf{99.81} & \textbf{98.20} \\ 
\midrule
\multirow{12}{*}{\rotatebox[origin=c]{90}{\shortstack{DynamiCrafter}}}
& RGB only & 30.49  & 94.31 & 85.90 \\
& LLL only & \textbf{33.23} & 50.93 & 51.78 \\
& HHH only & 33.19 & 51.29 & 51.21 \\
& LLH + RGB & 31.35 & 95.82 & 92.82 \\
& LHL + RGB & 31.12 & 95.56 & 91.17 \\
& HLL + RGB & 31.11 & 95.07 & 91.33 \\
& HHL + RGB & 30.98 & 95.28 & 90.50 \\
& HLH + RGB & 31.01 & 95.46 & 90.26 \\
& LHH + RGB & 30.55 & 95.33 & 89.21 \\
& HHH + RGB & 30.35  & 94.78 & 88.93 \\
\cmidrule{2-5}
& LLL + RGB \textbf{(Ours)} & 31.63  & \textbf{99.98} & \textbf{99.75} \\ 
\bottomrule
\end{tabular}
\end{table}

\vspace{0.5em}
\noindent \textbf{Watermark Decoder with 3D DWT.}
Temporal consistency is crucial in both video watermarking and generation. 
To enhance this, we use the 3D Wavelet Transform (3D DWT), which captures spatio-temporal features by decomposing videos into wavelet representations.
Table~\ref{tab:Ablation on the DWT} compares 3D Discrete Cosine Transform, 3D Fast Fourier Transform, 3D DWT, and 2D DWT. Our experiments show that 3D DWT outperforms other methods in bit accuracy and video quality due to its superior localization in time and frequency.
In contrast, 2D DWT, which only captures spatial information, results in a loss of temporal consistency, degrading video quality metrics such as tLP and FVD. 
Based on these findings, we use 3D DWT in our watermark decoder.

\vspace{0.5em}
\noindent \textbf{Watermark Decoder with Low-frequency.}
As mentioned in Section~\ref{subsec:Robust Video Watermark Decoder}, we utilize only the low-frequency subband ($LLL$) among the 3D wavelet subbands to be robust against compression. 
Table \ref{tab:Comparison of different domain types} shows that relying solely on the frequency domain yields high PSNR but results in poor bit accuracy. On the other hand, using only the spatial domain provides high bit accuracy but suffers from severe vulnerability to H.264 compression. Moreover, the results containing more high-frequency components show degraded performance under H.264 compression, since such components are suppressed during compression. To mitigate this, we explore combining each wavelet subband with RGB features. Among all results in Table \ref{tab:Comparison of different domain types}, integrating RGB features with the low-frequency subband ($LLL$) provides the best trade-off between visual quality and decoding robustness. Consequently, this design makes our method particularly robust to standard compression techniques, which preserve low-frequency components.

\vspace{0.5em}
\noindent \textbf{Comparison with Other Fusion Strategies.}
We adopt a cross-attention module to effectively fuse the spatio-temporal low-frequency features with spatial RGB features. Due to the different nature of these feature domains, simple concatenation is insufficient for effective fusion. To show the effectiveness of the cross-attention module, we compare our method with the concatenation approach for feature fusion in Table~\ref{tab:Concat}. The results show that the cross-attention module leads to improved video quality and bit accuracy. As shown in Table~\ref{tab:No H264}, the low-frequency subband exhibits significantly greater robustness to H.264 compression than the high-frequency subbands, effectively maintaining message decoding performance. Moreover, the H.264 network further improves decoding accuracy against H.264 compression.

\begin{table*}[t!]
\centering
\renewcommand{\arraystretch}{1}
\caption{Experiments on feature fusion methods (Concatenation and Cross-attention)}
\label{tab:Concat}
\begin{tabular}{cccccccccc}
\toprule
\multicolumn{1}{c}{} & \multicolumn{1}{c}{} & \multicolumn{5}{c}{Video Quality Metrics} & & \multicolumn{2}{c}{Bit Accuracy (\%) $\uparrow$} \\
\cmidrule{3-7} \cmidrule{9-10}
\multicolumn{1}{c}{} & \multicolumn{1}{c}{Feature Fusion} & PSNR $\uparrow$ & SSIM $\uparrow$ & LPIPS $\downarrow$ & tLP $\downarrow$ & FVD $\downarrow$ & & None & H.264 (CRF=24) \\ 
\midrule
\multirow{2}{*}{\begin{tabular}[c]{@{}c@{}}Open Sora\end{tabular}} 
& \multicolumn{1}{l}{Concatenation} & 31.24 & 0.917 & 0.109 & 0.232 & 86.25 && 90.48 & 78.91\\
\cmidrule{2-10}
& \multicolumn{1}{l}{Cross Attn \textbf{(Ours)}} & \textbf{31.74} & \textbf{0.918} & \textbf{0.105} & \textbf{0.229} & \textbf{83.98} && \textbf{99.81} & \textbf{98.20} \\
\midrule
\multirow{2}{*}{\begin{tabular}[c]{@{}c@{}}DynamiCrafter\end{tabular}} 
& \multicolumn{1}{l}{Concatenation} & 31.57 & 0.912 & 0.069 & 0.193 & 71.48 && 92.08 & 80.26  \\
\cmidrule{2-10}
    & \multicolumn{1}{l}{Cross Attn \textbf{(Ours)}} & \textbf{31.63} & \textbf{0.914} & \textbf{0.066} & \textbf{0.189} & \textbf{68.94} && \textbf{99.98} & \textbf{99.75}\\
\bottomrule
\end{tabular}
\end{table*}

\begin{table*}[t!]
\vspace{-1em}
\centering
 \caption{Extreme H.264 results for Open-Sora~\cite{opensora} and DynamiCrafter~\cite{xing2025dynamicrafter}}
\renewcommand{\arraystretch}{1}
\setlength{\tabcolsep}{2pt} 
\begin{tabular}{lccccccccccccccc}
\toprule
\multicolumn{1}{c}{} & \multicolumn{7}{c}{Open Sora} & \multicolumn{1}{c}{} & \multicolumn{7}{c}{DynamiCrafter} \\ 
\cmidrule{2-8} \cmidrule{10-16}
Method & None &CRF=25 & CRF=26 & CRF=27 & CRF=28 & CRF=29 & CRF=30 && None & CRF=25 & CRF=26 & CRF=27 & CRF=28 & CRF=29 & CRF=30 \\ 
\midrule
\multirow{1}{*}VideoShield~\cite{hu2025videoshield} & 99.52 & 96.87 & 96.54 & 96.07 & 95.67 & 94.54 & 93.44 && 95.40 & 94.53 & 94.21 & 94.01 & 93.84 & 93.56 & 92.42
\\
\multirow{1}{*}VideoMark~\cite{hu2025videomark} & 99.45 & 95.38 & 95.21 & 95.03 & 94.24 & 94.07 & 93.38 && 95.13 & 94.52 & 93.67 & 93.24 & 92.76 & 92.33 & 92.04
\\
\cmidrule{1-16}
    \multirow{1}{*}{{LVMark}~(STE)} & 97.92 & 96.54 & 96.34 & 96.21 & 96.09 & 95.87 & 95.54 && 98.78 & 97.13 & 97.04 & 96.95 & 96.52 & 96.43 & 96.13 \\ 
    \multirow{1}{*}{\textbf{LVMark}~\textbf{(Ours)}} & \textbf{99.81} & \textbf{98.92} & \textbf{98.53} & \textbf{98.21} & \textbf{97.94} & \textbf{97.82} & \textbf{97.32} && \textbf{99.98} &  \textbf{99.52} & \textbf{99.42} & \textbf{99.28} & \textbf{98.99} & \textbf{98.49} & \textbf{98.05} \\ 
\bottomrule
\end{tabular}
\label{tab:More H264 Results}
\end{table*}

\begin{table}[t!]
\vspace{-1em}
\centering
\setlength{\tabcolsep}{11.8pt} 
\renewcommand{\arraystretch}{1}
\caption{Performance of the H.264 network}
\label{tab:h264 network}
\begin{tabular}{lccc}
\toprule
Dataset & PSNR $\uparrow$ & SSIM $\uparrow$ & LPIPS $\downarrow$ \\ 
\midrule
Panda70M~\cite{chen2024panda} & 36.41 & 0.965 & 0.048 \\
\bottomrule
\end{tabular}
\vspace{-1em}
\end{table}

\vspace{0.5em}
\noindent \textbf{Performance of H.264 network.}
We evaluate our H.264 network with randomly selected 2K test videos from Panda-70M dataset.
According to Table~\ref{tab:h264 network}, our H.264 network generates videos that are highly similar to those compressed with actual H.264, allowing us to replace real H.264 compression with a pre-trained H.264 network.

\begin{figure}[t!]
\vspace{-1em}
    \begin{center}
        \includegraphics[width=1\linewidth]{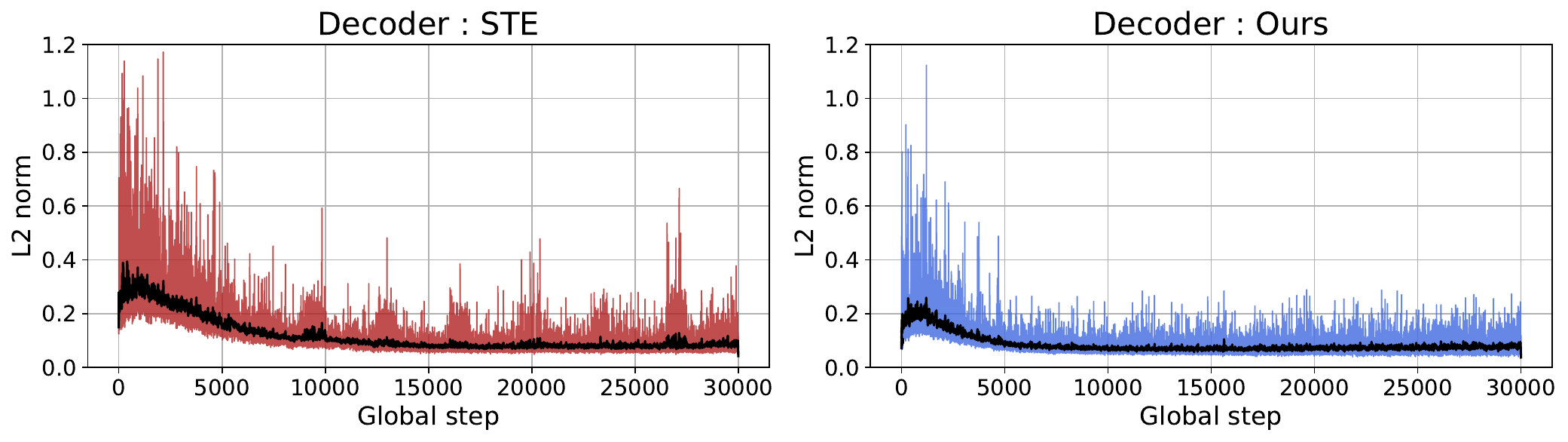}
    \end{center}
    \vspace{-1.5em}
    \caption{Gradient stability comparison between the Straight-Through Estimator and our H.264 network. The gradient L2 norm of the decoder is plotted over global training steps. The black line represents the moving average.}
    \label{fig:gradient}
\end{figure}

\vspace{0.5em}
\noindent \textbf{Comparison with Straight-Through Estimator.}
Since H.264 compression involves non-differentiable operations, it cannot be directly incorporated into gradient-based optimization. An approach to bypass such non-differentiability is utilizing the Straight-Through Estimator (STE)~\cite{bengio2013estimating}, which approximates the gradient of the non-differentiable operation using the identity function. Instead of relying on such approximations, we employ a differentiable proxy network to approximate the behavior of H.264 compression during training.
To evaluate the effectiveness of our approach, we compare our H.264 network with the STE-based H.264. As shown in Figure~\ref{fig:gradient}, the STE-based H.264 exhibits significant fluctuations and high variance in its gradients. This instability originates from the fact that the gradients produced by STE are coarse approximations rather than the true gradients of the loss function. In contrast, our H.264 network produces significantly smoother gradients with substantially reduced variance. The moving-average trend shows more stable optimization behavior, suggesting that the differentiable proxy network enables more reliable gradient propagation during training. This improved gradient stability contributes to more robust watermark decoding performance.

\begin{table}[t!]
\centering
\renewcommand{\arraystretch}{1.3} 
\caption{Other Codec results for Open-Sora~\cite{opensora} and DynamiCrafter~\cite{xing2025dynamicrafter}}
\label{tab:Other Codec}
\begin{tabular}{llccccc}
\toprule
\multicolumn{2}{c}{} & \multicolumn{5}{c}{Bit Accuracy (\%)~$\uparrow$} \\
\cmidrule{3-7} 
& Method & None & H.264  & H.265 & MPEG-4  & VP9 \\ 
\midrule
\multirow{4}{*}{\rotatebox[origin=c]{90}{Open Sora}}
& VideoShield~\cite{hu2025videoshield} & 99.52 & 96.87 & 95.81 & 95.17 & 83.49 \\
& VideoMark~\cite{hu2025videomark} & 99.45 & 95.38 & 94.28 & 92.09 & 57.31  \\
\cmidrule{2-7}
& LVMark~(STE) & 97.92 & 96.54 & 96.42 & 95.92 & 94.88  \\ 
& \textbf{LVMark}~\textbf{(Ours)} & \textbf{99.81} & \textbf{97.32} & \textbf{97.04} & \textbf{96.87} & \textbf{96.64}    \\ 
\midrule
\multirow{4}{*}{\rotatebox[origin=c]{90}{\shortstack{DynamiCrafter}}}
& VideoShield~\cite{hu2025videoshield} & 95.40 & 94.53 & 94.19 & 93.63 & 81.60  \\
& VideoMark~\cite{hu2025videomark} & 95.13 & 94.52 & 93.25 & 93.33 & 53.78  \\
\cmidrule{2-7}
& LVMark~(STE) & 98.78 & 97.13  & 96.44  & 94.51 & 95.29  \\ 
& \textbf{LVMark}~\textbf{(Ours)} & \textbf{99.98} & \textbf{98.05} & \textbf{97.89} & \textbf{97.79} & \textbf{96.12}    \\ 
\bottomrule
\end{tabular}
\end{table}

\vspace{0.5em}
\noindent \textbf{Robustness under extreme H.264 compression.}
To investigate robustness against severe compression, we evaluate our method using 512-bit messages under H.264 compression with extreme Constant Rate Factors (CRF=25 to 30). As detailed in Table~\ref{tab:More H264 Results}, our method outperforms other methods across all compression levels. Specifically, on Open-Sora~\cite{opensora}, as compression reaches the extreme CRF=30, the bit accuracy of VideoMark~\cite{hu2025videomark} and VideoShield~\cite{hu2025videoshield} suffer severe drops of 6.07\%p  and 6.08\%p, respectively. In contrast, our method exhibits a much lower performance degradation, dropping by only 2.49\%p while maintaining a high accuracy of 97.32\%. Similarly, on DynamiCrafter~\cite{xing2025dynamicrafter}, our method drops merely 1.93\%p. Furthermore, our method consistently surpasses the STE, confirming the effectiveness of our method. These minimal degradation rates show that our method is highly practical and robust for real-world environments.

\vspace{0.5em}
\noindent \textbf{Robustness Under Other Codecs.}
To comprehensively evaluate the robustness of our method against various compression standards, we conduct experiments using 512-bit messages across different codecs with a CRF is 25. While H.264 is the most widely adopted standard, real-world applications employ alternative codecs such as H.265 for superior compression ratios, MPEG-4, and VP9, which is utilized in streaming environments. As shown in Table~\ref{tab:Other Codec}, recent video diffusion watermarking methods exhibit severe vulnerability to alternative compression schemes, with their bit accuracy dropping significantly under VP9 compression. Specifically, the bit accuracy of VideoMark~\cite{hu2025videomark} drops significantly by 42.14\%p on Open-Sora~\cite{opensora} and 41.35\%p on DynamiCrafter~\cite{xing2025dynamicrafter}. Similarly, VideoShield~\cite{hu2025videoshield} shows substantial drops of 16.03\%p and 13.80\%p in bit accuracy on Open-Sora and DynamiCrafter~\cite{xing2025dynamicrafter}, respectively.
In contrast, our method shows generalizability, consistently maintaining a high bit accuracy of over 96\% across all evaluated codecs. This indicates our spatio-frequency fusion is inherently robust to common quantization processes across various codecs.

\begin{table}[t!]
\centering
\setlength{\tabcolsep}{5pt} 
\renewcommand{\arraystretch}{2} 
\caption{Ablation on weighted patch loss}
\label{tab:Ablation on the patch loss}
\begin{tabular}{llccc}
\toprule
& Loss Type  & Bits Acc. $\uparrow$ & PSNR $\uparrow$ & LPIPS $\downarrow$  \\ 
\midrule
\multirow{3}{*}{\rotatebox[origin=c]{90}{Open Sora}}
& VGG Loss & 99.78 & 30.86 & 0.107 \\
& VGG + MAE Loss & 99.80 & 31.12 & 0.139  \\
& VGG + Patch Loss \textbf{(Ours)}& \textbf{99.81} & \textbf{31.74} & \textbf{0.105}    \\ 
\midrule
\multirow{3}{*}{\rotatebox[origin=c]{90}{\shortstack{DynamiCrafter}}}
& VGG Loss & 99.86 & 30.74 & 0.071 \\
& VGG + MAE Loss & 99.82 & 31.03 & 0.107  \\
& VGG + Patch Loss \textbf{(Ours)}& \textbf{99.98} & \textbf{31.63} & \textbf{0.066}    \\ 
\bottomrule
\end{tabular}
\vspace{-0.5em}
\end{table}

\begin{figure}[t!]
    \begin{center}
        \includegraphics[width=1\linewidth]{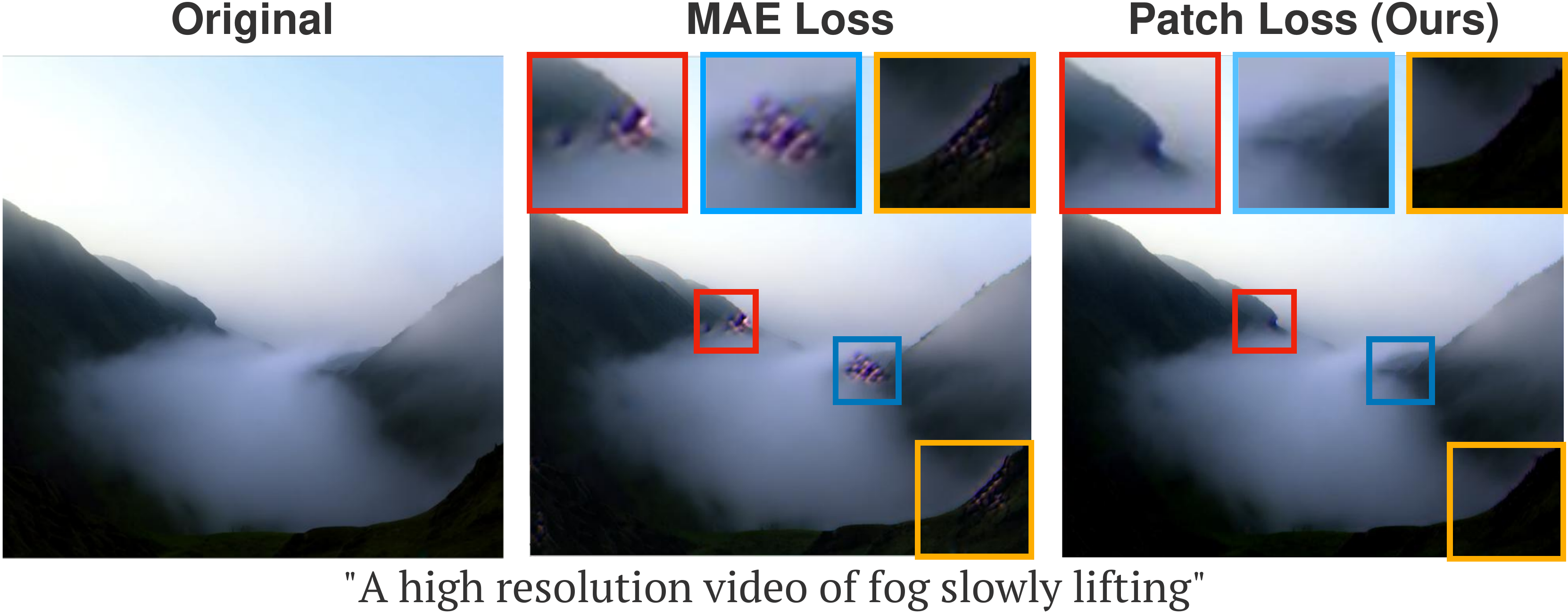}
    \end{center}
    \vspace{-1.5em}
    \caption{Visual impact of weighted patch loss. We visualize a video frame trained with and without weighted patch loss. We crop the local regions containing artifacts. Both results used VGG Loss~\cite{czolbe2020loss} with the default setting.}
    \label{fig:patch loss}
    \vspace{-1em}
\end{figure}

\begin{table}[t!]
\centering
\caption{Impact of patch size}
\renewcommand{\arraystretch}{1.4}
\begin{tabular}{clcc}
\toprule
\multicolumn{1}{c}{} & \multicolumn{1}{c}{Patch size} & \multicolumn{1}{c}{LPIPS $\downarrow$} & \multicolumn{1}{c}{Bits Accuracy (\%) $\uparrow$} \\ 
\midrule
\multirow{4}{*}{\rotatebox[origin=c]{90}{Open Sora}}
& 4 & 0.121 & 99.75 \\
& 8 \textbf{(Ours)} & \textbf{0.105} & \textbf{99.81} \\
& 16 & 0.112 & 99.74 \\
& 32 & 0.115 & 99.78 \\ 
\midrule
\multirow{4}{*}{\rotatebox[origin=c]{90}{\shortstack{DynamiCrafter}}}
& 4 & 0.099 & 99.97 \\
& 8 \textbf{(Ours)} & \textbf{0.066} & \textbf{99.98} \\
& 16 & 0.077 & 99.96 \\
& 32 & 0.082 & 99.97 \\ 
\bottomrule
\end{tabular}
\vspace{0em}
\label{tab:Patch Loss Size}
\end{table}

\vspace{0.5em}
\noindent \textbf{Impact of Weighted Patch Loss.}
As described in Section~\ref{subsec:Training Objectives}, weighted patch loss mitigates localized artifacts, enhancing video quality.
VGG Loss~\cite{czolbe2020loss} is often used to improve the perceptual quality of images, but it has limitations in enhancing PSNR, which measures pixel-wise differences. 
To address this, combining VGG Loss with Mean Absolute Error (MAE) Loss, which focuses on pixel-wise differences, can be effective. 
However, as shown in Figure~\ref{fig:patch loss}, our experiment reveals that using both losses together introduces localized artifacts, as the embedded watermark in the video creates artifacts that degrade visual quality.
Table~\ref{tab:Ablation on the patch loss} shows that weighted patch loss enables higher bit accuracy while preserving video quality, outperforming MAE Loss.
This highlights the effectiveness of weighted Patch Loss, considering the trade-off between bit accuracy and visual quality.
As indicated in Table~\ref{tab:Patch Loss Size}, we further evaluate the impact of various patch sizes. The results indicate that a patch size of 8 provides the best balance between bit accuracy and visual quality, yielding the lowest LPIPS scores and the highest bit accuracy. When the patch size is too small, the loss overly emphasizes localized variations, resulting in reduced visual coherence. Conversely, larger patch sizes tend to over-smooth local error signals, limiting fine-grained control and degrading visual quality. Therefore, we use a patch size of 8, which makes \(32^{2}\) patches on 256$\times$256 resolution, as the default setting in our experiments.

\vspace{2em}
\noindent \textbf{Error Correction Coding.}
Watermarking methods often utilize error correction coding to recover from bit errors~\cite{cika2007new, shiddik2019compressive, wang2024robust, wu2010bch, usman2010bch}. In this work, we adopt Bose–Chaudhuri–Hocquenghem (BCH) code, which is often used to encode watermarks during the generation of the watermarked video and decode them during message retrieval to detect and correct bit errors. BCH code is a cyclic error-correcting code that is capable of detecting and correcting multiple bit errors. It is constructed using a generator polynomial over a finite field, which systematically encodes messages to enhance their resilience to errors. 
By storing each model user’s watermark key and BCH-encoded key in a database, our method enables the identification of video ownership. To ensure a fair comparison, we apply the same BCH code parameters across all evaluated baseline methods. 
The BCH code is defined as \( \text{BCH}(k, t) \), where \( k \) is the message length and \( t \) is the error correction capacity. For different capacities, we use the following parameters: BCH(32,1) for 32-bit, BCH(96,2) for 96-bit, and BCH(512,14) for 512-bit.
This setting enables error correction with an accuracy of 97-98\% across all capacities. 
Table~\ref{tab:effectiveness of ECC} shows that the bit accuracy before applying BCH in the 512-bit capacity remains sufficiently high.

\begin{table}[t!]
\centering
\setlength{\tabcolsep}{2em} 
\caption{Effectiveness of Error Correction Coding (ECC)}
\label{tab:effectiveness of ECC}
\begin{tabular}{ccc}
\midrule
 & W/o ECC & W/ ECC \\ 
\midrule
Open Sora & 98.21 & 99.81 \\
DynamiCrafter & 98.87 & 99.98 \\
\midrule
\end{tabular}
\end{table}

\begin{figure}[t!]
    \begin{center}
        \includegraphics[width=1\linewidth]{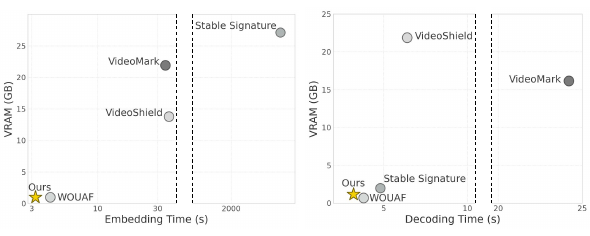}
    \end{center}
    \vspace{-1.5em}
    \caption{Comparison of VRAM footprint and processing time. Our method requires significantly less memory while maintaining highly efficient embedding and decoding speeds compared to the other methods.}
    \label{fig:vram}
    \vspace{-1em}
\end{figure}

\vspace{0.5em}
\noindent \textbf{Computational Cost.}
Figure~\ref{fig:vram} and Table~\ref{tab:Cost} present the computational efficiency and memory footprint of our method, compared to the others. As shown in Figure~\ref{fig:vram}, our method requires substantially less VRAM while maintaining faster embedding and decoding times compared to the other methods such as VideoShield~\cite{hu2025videoshield} and VideoMark~\cite{hu2025videomark}. Since these methods rely on diffusion inversion, they require substantial GPU memory during watermark extraction. Additionally, Stable Signature~\cite{fernandez2023stable} requires a separate and complete training process for each new watermark. In contrast, the weight modulation approach allows us to embed random, multiple messages into the video diffusion model with only a single training process. Moreover, unlike WOUAF~\cite{kim2024wouaf}, which modifies all layers of the model, our method selectively modulates layers, leading to more efficient and faster watermark embedding.
Table~\ref{tab:Cost} demonstrates that the proposed framework introduces only minimal computational overhead. Under FP32 precision, the memory footprint during generation  increases by 0.35 GB (from 26.30 GB to 26.65 GB), and the generation runtime remains virtually identical, adding only 0.04 seconds. 
Moreover, adopting FP16 substantially alleviates the computational burden, drastically reducing the VRAM requirements for training, generation, and decoding. Despite this massive reduction in computational cost, our method securely retains its core capabilities, maintaining the performance. These results show that our method provides robust watermark embedding while preserving computational efficiency and visual fidelity.

\vspace{0.5em}
\noindent \textbf{Scalable Online Serving}.
To support scalable per-user watermarking in a user-specific ownership scenario, our method inherently prevents race conditions among concurrent users by keeping the deployed base model weights strictly frozen. Rather than performing in-place updates on the shared global weights, the mapping network dynamically processes each incoming user message to generate modulated weights on the fly. Since these user-specific modulations are instantiated and utilized exclusively for the corresponding generation process, multiple users can simultaneously generate uniquely watermarked videos without modifying the shared base model.

\begin{table}[t!]
\centering
\setlength{\tabcolsep}{1em} 
\caption{Comparison of Computational Costs}
\begin{tabular}{@{}lcc@{}}
\toprule
Metric & \begin{tabular}[c]{c}Original \\ FP32 / FP16\end{tabular} & \begin{tabular}[c]{c}Watermarked \\ FP32 / FP16\end{tabular} \\ \midrule
Train (GB) & - & 25.12 / 15.89 \\
Generation (GB) & 26.30 / 21.18 & 26.65 / 21.47 \\
Runtime (s, Generation) & 31.94 / 31.91 & 31.98 / 31.96 \\
Bits Accuracy &- &99.98 / 99.96 \\
Combine Attack &- &99.16 / 99.15 \\
PSNR &-& 31.63 / 31.60 \\
SSIM &-& 31.63 / 31.60 \\
LPIPS &-& 31.63 / 31.60 \\
tLP &-& 0.189 / 0.189 \\
FVD &-& 68.94 / 68.92 \\
\bottomrule
\label{tab:Cost}
\end{tabular}
\end{table}

\section{Conclusion}
We proposed LVMark, a robust watermarking method for video diffusion models that can be applied to both U-Net and DiT-based architectures. 
Our method captures spatio-temporal information by effectively fusing the 3D wavelet domain and the RGB spatial domain for accurate message extraction from generated videos, even in the case of video and model modifications.
Additionally, our method generates high-quality watermarked videos by selectively modulating latent decoder weights according to the importance of each layer.
By balancing the trade-off between video quality and bit accuracy with tailored objective functions, including weighted patch loss, LVMark successfully embeds watermarks into video diffusion models.
Experimental results show that our method is a practical video diffusion watermarking approach, achieving a high capacity of 512-bit while preserving original content and ensuring robust watermark retention against various attacks, including model-based ones.

\vspace{0.5em}
\noindent \textbf{Limitations \& Future Work.} Our method excels in watermarking video diffusion models.
However, training requires around 25GB memory in 32-bit floating point precision, due to the large size of the diffusion models and videos. 
Future work could explore memory-efficient watermarking techniques.

\section*{Acknowledgement}
This work was supported by Culture, Sports and Tourism R\&D Program through the Korea Creative Content Agency grant funded by the Ministry of Culture, Sports and Tourism (International Collaborative Research and Global Talent Development for the Development of Copyright Management and Protection Technologies for Generative AI, RS-2024-00345025, 37\%; Research on neural watermark technology for copyright protection of generative AI 3D content, RS-2024-00348469, 25\%), the National Research Foundation of Korea(NRF) grant funded by the Korea government(MSIT) (RS-2025-00521602, 35\%), Institute of Information \& communications Technology Planning \& Evaluation (IITP) \& ITRC(Information Technology Research Center) grant funded by the Korea government(MSIT) (No.RS-2019-II190079, Artificial Intelligence Graduate School Program(Korea University), 1\%; IITP-2025-RS-2024-00436857, 1\%), and the Advanced GPU Utilization Support Program funded by the Government of the Republic of Korea (Ministry of Science and ICT).

\begin{figure*}[t!]
    \begin{center}
        \includegraphics[width=0.6\textwidth ]{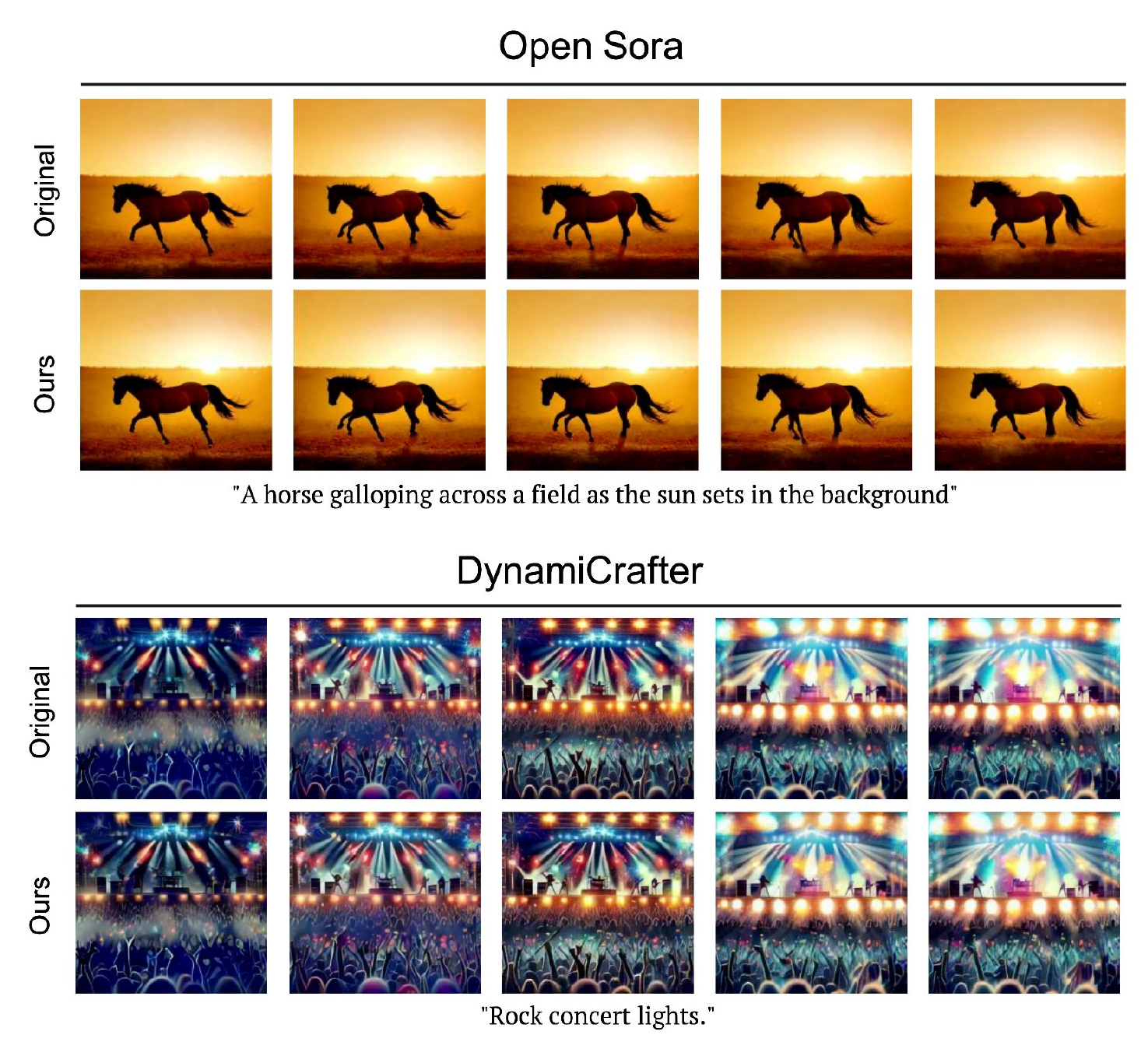}
    \end{center}
    \vspace{-2em}
    \caption{Visualization of generated videos. This figure illustrates the frames of videos generated by Open-Sora~\cite{opensora} and DynamiCrafter~\cite{xing2025dynamicrafter}.}
    \label{fig:further_qualitative}
\end{figure*}

\vfill

\bibliographystyle{IEEEtran} 
\bibliography{IEEEfull}

\vspace{-4em}
\begin{IEEEbiography}[{
\includegraphics[width=1in,height=1.25in,clip,keepaspectratio]{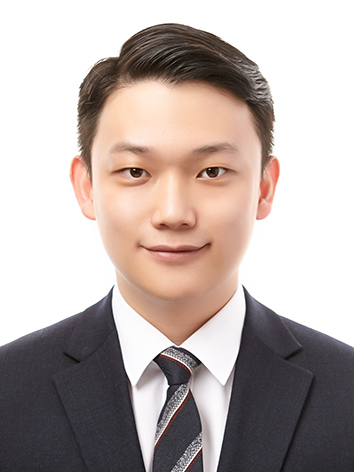}}]{Youngdong Jang}
received the B.S. degree in mathematics from Dongguk University, Seoul, Republic of Korea, in 2021. He is currently pursuing the Ph.D. degree in artificial intelligence at Korea University, Seoul, Republic of Korea. His research expertise lies in computer vision, computer graphics, image/video processing, and AI safety. Specifically, his work focuses on robust watermarking techniques for copyright protection in AI models, 3D/4D scene understanding.
\vspace{-4em}
\end{IEEEbiography}

\begin{IEEEbiography}[{\includegraphics[width=1in,height=1.25in,clip,keepaspectratio]{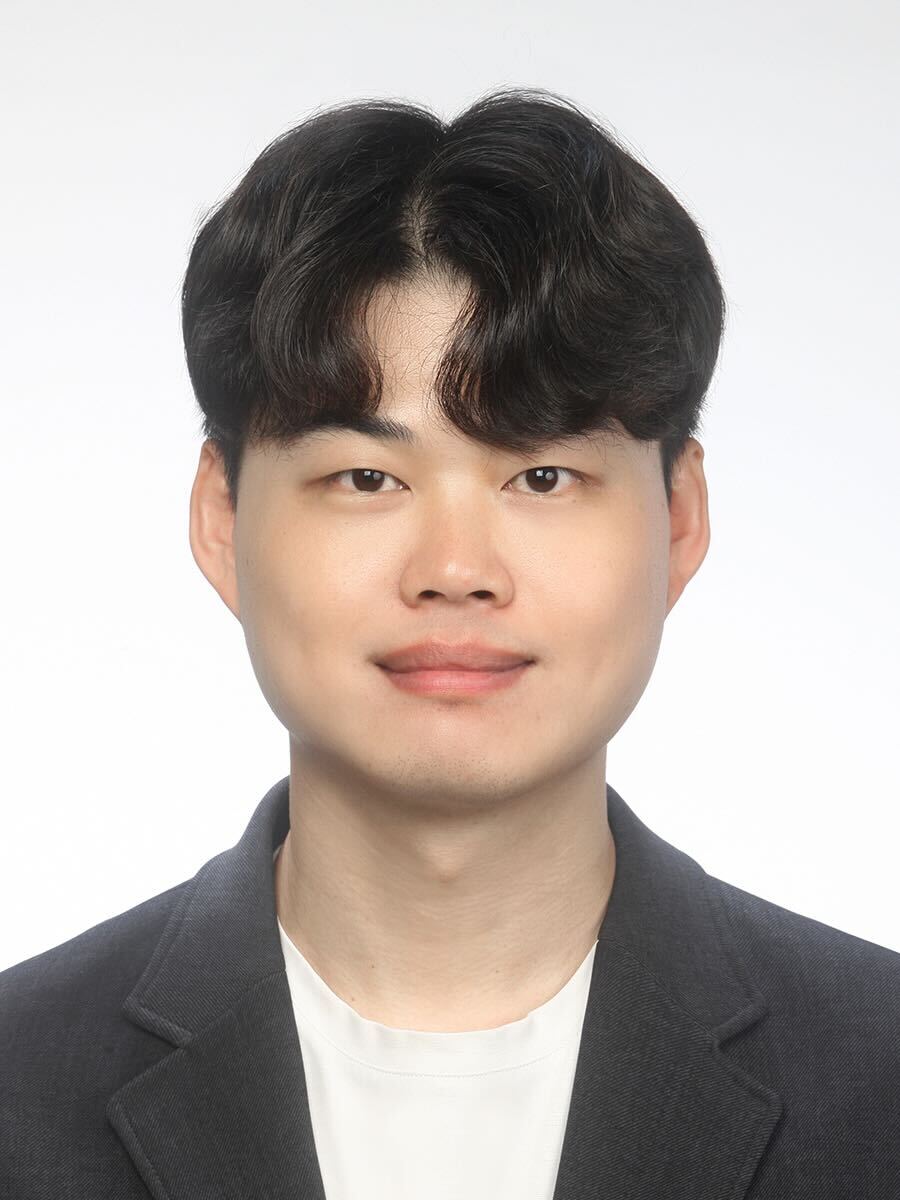}}]{MinHyuk Jang} received the B.S. degree in Computer Science from Kookmin University and the M.S. degree from the Department of Artificial Intelligence at Korea University. As an AI researcher, his work focuses on advancing generative AI, with a particular emphasis on 3D-aware structural reconstruction, pose-driven image-to-video (I2V) generation, and digital human animation. He is deeply interested in developing efficient frameworks that bridge the gap between 2D visual synthesis and 3D spatial understanding, aiming to create highly realistic and temporally coherent digital content for next-generation immersive platforms. Additionally, his research extends to AI security, specifically focusing on robust watermarking technologies to ensure the integrity and secure use of generative models and AI-generated media.
\vspace{-4em}
\end{IEEEbiography}

\begin{IEEEbiography}[{\includegraphics[width=1in,height=1.25in,clip,keepaspectratio]{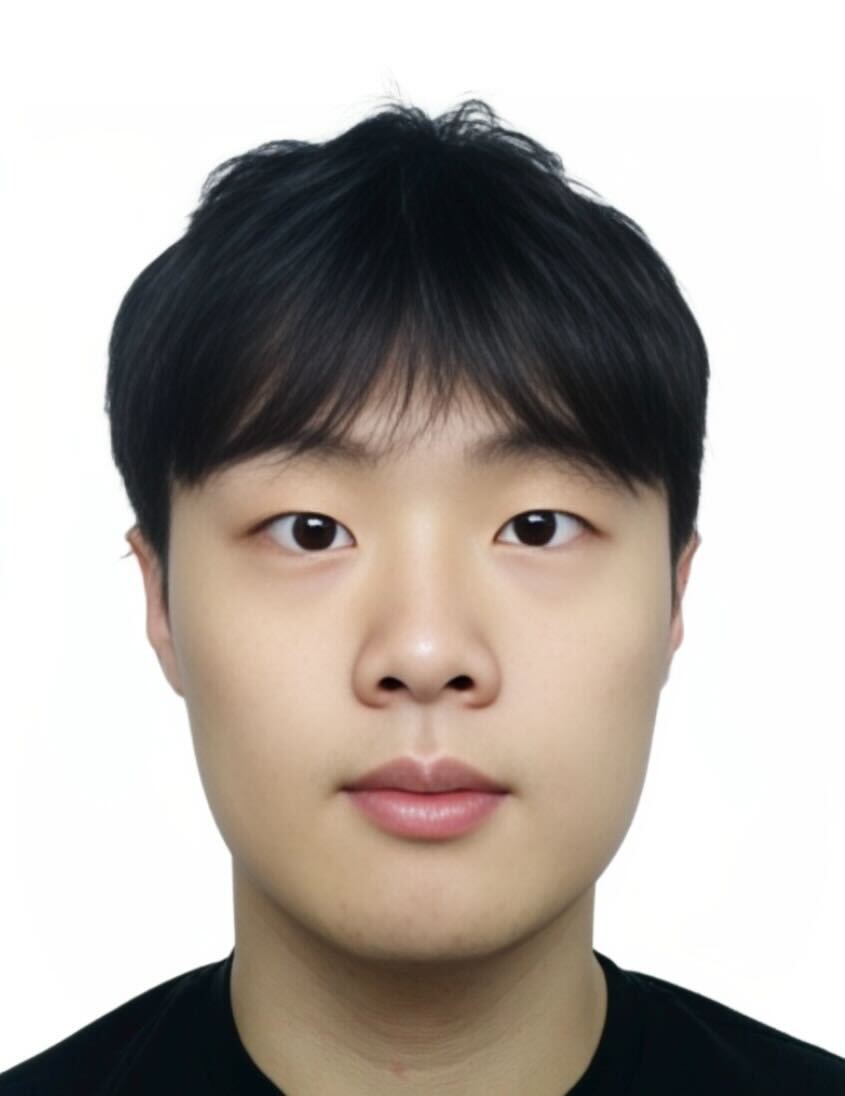}}]{JaeHyeok Lee} received his B.S. degree from Hongik University, Seoul, and is currently pursuing an integrated M.S./Ph.D. degree at Korea University, starting in 2025. His primary research focuses on Large Language Model (LLM) agents and Human-Scene Interaction (HSI), with a particular emphasis on the development of self-evolving agentic AI.
\vspace{-4em}
\end{IEEEbiography}

\begin{IEEEbiography}[{\includegraphics[width=1in,height=1.25in,clip,keepaspectratio]{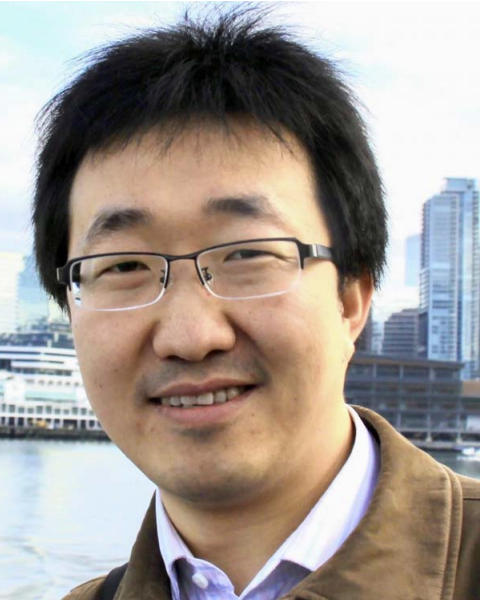}}]{Feng Yang}
(Senior Member, IEEE) received the B.Eng. and M.Eng. degrees in automatic control from Tsinghua University, Beijing, China, in 2004 and 2007, respectively, and the Ph.D. degree in communication systems from the École Polytechnique Fédérale de Lausanne (EPFL), Lausanne, Switzerland, in 2012. He was a Research Assistant with Broadband Network and Digital Multimedia Laboratory, Tsinghua University, and Audiovisual Communications Laboratory, EPFL. He interned at Intel China Research Center, Beijing, and Nokia Research Center, Palo Alto, CA, USA. He was a
Postdoctoral Researcher at Illumination and Imaging Laboratory, The Robotics Institute, Carnegie Mellon University (CMU), Pittsburgh, PA, USA. He was a Tech Lead Manager and a Senior Staff Software Engineer at Google Research, Mountain View, CA, USA. He is currently a Principal Scientist (Director) at GenAI, Google DeepMind, Mountain View, CA, USA, leading a team working on research and productionization of Gemini, Imagen and Veo. His research interests include LLM\&VLM, multimodal understanding \& generation, responsible AI, computer vision, multimedia processing and communications.
\vspace{-4em}
\end{IEEEbiography}

\begin{IEEEbiography}[{\includegraphics[width=1in,height=1.25in,clip,keepaspectratio]{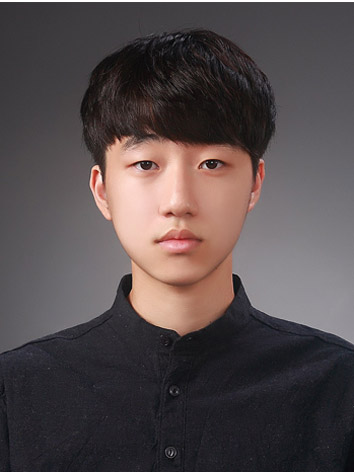}}]{Gyeongrok Oh}
received his B.S. degree in Computer Science and Engineering from Inha University in 2022. He is currently a graduate student at Department of Aritificial Intelligence at the Korea University. His research interests include multi-modal representation, image restoration.
\vspace{-4em}
\end{IEEEbiography}

\begin{IEEEbiography}[{\includegraphics[width=1in,height=1.25in,clip,keepaspectratio]{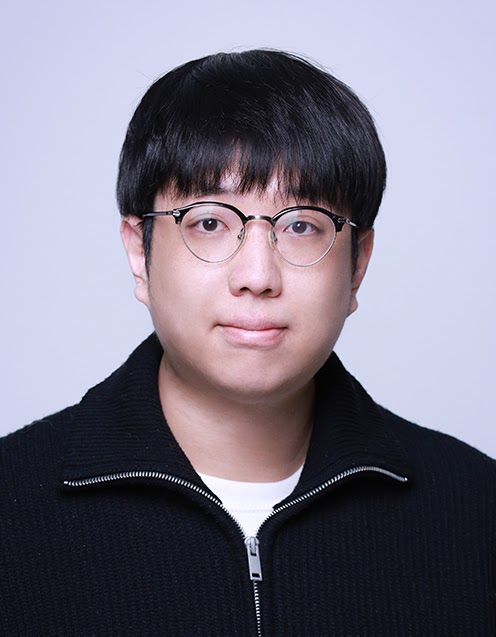}}]{Jongheon Jeong} is an assistant professor in the Department of Artificial Intelligence at Korea University (KU), leading the Trustworthy AI Lab. He obtained his Ph.D. in Electrical Engineering from KAIST in 2023, and B.S. in Mathematics and Computer Science from KAIST in 2017. During his Ph.D. studies, he worked at Amazon Web Services (AWS) as an Applied Scientist Intern in 2021 (Seattle, WA) and 2022 (Bellevue, WA). He received the Best Doctoral Dissertation Award from the KAIST College of Engineering in 2024, and the Qualcomm Innovation Fellowship Korea 2020 from two of his papers. His research interest lies in broad areas of AI Safety, towards making deep learning more reliable against various unseen scenarios.
\vspace{-4em}
\end{IEEEbiography}

\begin{IEEEbiography}[{\includegraphics[width=1in,height=1.25in,clip,keepaspectratio]{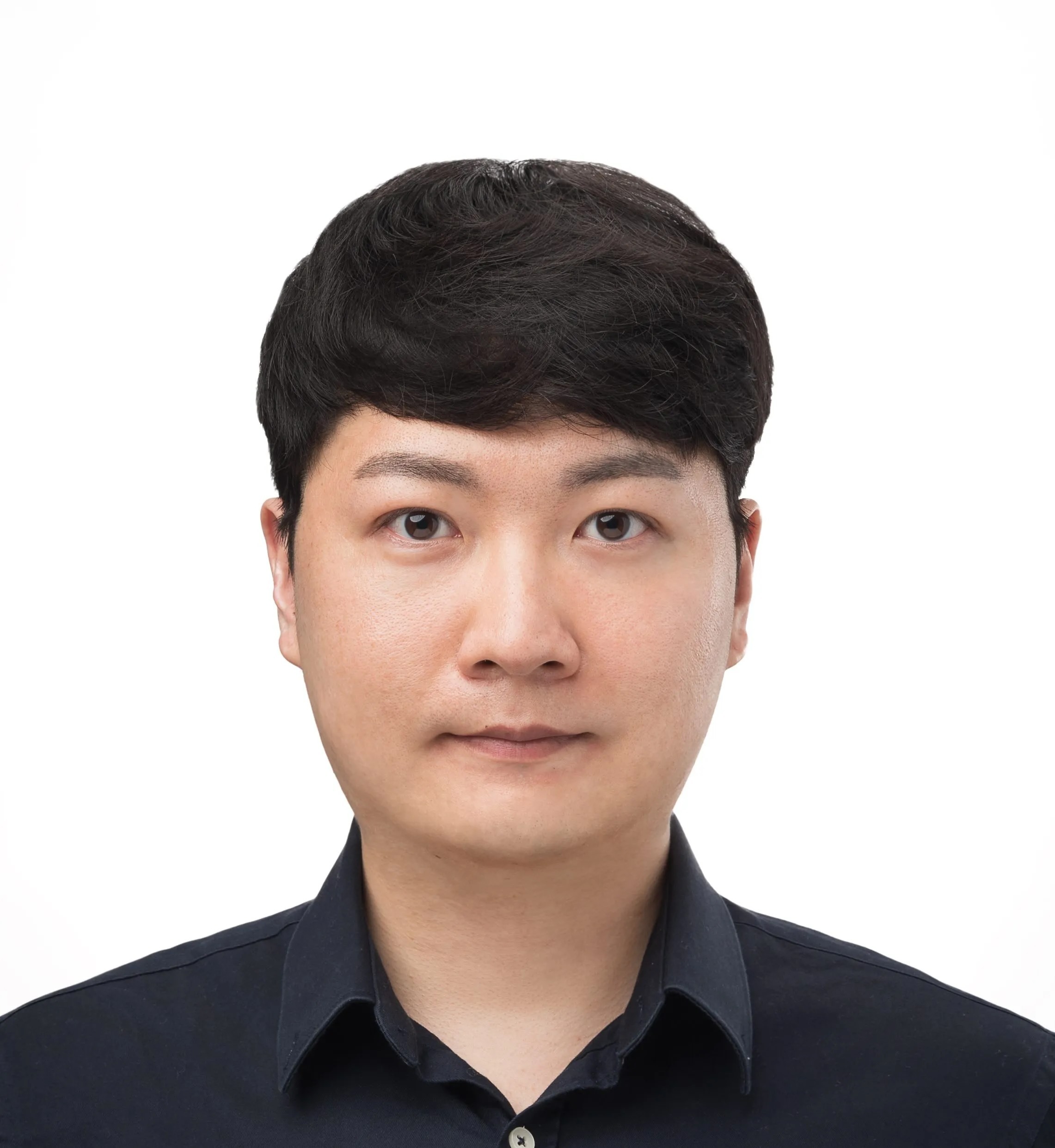}}]{Sangpil Kim} is an Associate Professor in the Department of Artificial Intelligence at Korea University. He received his Ph.D. in Electrical and Computer Engineering from Purdue University and his B.Sc. in Computer Science from Korea University. His research focuses on computer vision and artificial intelligence.
\end{IEEEbiography}

\end{document}